%% file: main.tex
\pdfoutput=1
\documentclass{article}

\PassOptionsToPackage{numbers,compress}{natbib}

\usepackage[eandd, final]{neurips_2026}  

\usepackage[utf8]{inputenc} 
\usepackage[T1]{fontenc}    
\usepackage[hypertexnames=false]{hyperref}       
\hypersetup{colorlinks=true, citecolor=blue, linkcolor=blue, urlcolor=blue}
\usepackage{url}            
\usepackage{booktabs}       
\usepackage{amsfonts}       
\usepackage{nicefrac}       
\usepackage{microtype}      
\usepackage{xcolor}         
\usepackage{graphicx}       
\usepackage{amsmath}
\usepackage{wrapfig}
\usepackage{float}

\providecommand{\FloatBarrier}{\clearpage}

\title{More Is Not More: What Matters for\texorpdfstring{\\}{ } Opinion Diversity in LLM Simulations?}

\author{%
  Qiyang Yao\thanks{Now at New York University.} \\
  Independent Researcher \\
  \texttt{qy1138@nyu.edu}
}

\begin{document}

\maketitle

\begin{abstract}
Large language models are increasingly used to simulate diverse human opinions in open-ended tasks such as synthetic surveys, focus group modeling, and public opinion prediction. However, LLM outputs exhibit systematic opinion homogenization, and a simulation cannot reproduce a population's opinion distribution without first covering its range of viewpoints. Practitioners have explored various interventions to increase diversity, but the landscape remains fragmented: different methods are evaluated in isolation with incomparable metrics, and in practice they are typically deployed and upgraded simultaneously, making it difficult to attribute gains to specific components. To advance a more scientific understanding of LLM output diversity, we design a factorial experiment that separates two primary intervention dimensions: input conditioning (operationalized through persona depth) and interaction architecture. We evaluate all conditions on 100 real-user open-ended questions across 7 models, measuring diversity with multiple complementary metrics. Our findings challenge several common assumptions. First, more persona detail does not monotonically increase diversity. The initial step of persona conditioning already captures the majority of the gain, while further elaboration with demographic detail does not consistently improve and can reduce diversity on some models; a blinded human evaluation of matched Role and Basic responses points in the same direction. Second, rather than seeking a single best interaction architecture, we find that different architectures explore largely non-overlapping opinion regions. Combining multiple architectures yields broader coverage than optimizing any one. Third, commonly attempted low-cost alternatives such as raising sampling temperature and adding diversity instructions produce negligible effects compared to structured interventions. Overall, our results indicate that, within the tested settings, diversity does not scale along any single dimension but is sensitive to the structural form and combination of interventions. The field needs to move toward empirically grounded diversity strategy design over assumption-driven exploration.
\end{abstract}

\section{Introduction}
\label{sec:intro}

Large language models excel at reasoning tasks with determinate answers, but a growing body of research and applications directs these models toward a different class of problems: generating synthetic survey responses~\citep{argyle2023,bisbee2024}, moderating focus group discussions~\citep{focusagent2024}, modeling public opinion distributions~\citep{santurkar2023,durmus2023}, and predicting social behavior~\citep{horton2023,park2023generative,hewitt2024predicting}.
In these open-ended settings, value derives not from the correctness of any single response but from whether the collective output covers a heterogeneous, distinguishable set of human stances, judgments, and rationales. A synthetic focus group is of little use if every participant expresses the same concern, reaches the same conclusion, and differs only in wording. This makes broad opinion coverage a core requirement---a prerequisite for, though not a guarantee of, faithful simulation.

Yet LLM opinion diversity faces systematic limitations. Different models converge toward similar viewpoints~\citep{park2024diminished, jiang2025}, alignment training substantially narrows the opinion distribution~\citep{kirk2024,santurkar2023}, and LLM-simulated survey responses underestimate the variance observed in human populations, even when conditioned on demographic profiles~\citep{bisbee2024}.

A growing body of work proposes interventions to recover this lost diversity. These range from input-level strategies---persona prompting~\citep{hu2024}, explicit diversity instructions~\citep{hayati2024}, multilingual prompting~\citep{wang2025multilingual}---to architectural interventions such as multi-agent debate~\citep{liang2024,huzhe2025debate} and structured discussion protocols~\citep{chen2024}, as well as decoding-time techniques including verbalized sampling~\citep{zhang2025verbalized} and CoT-based elicitation~\citep{meincke2024}. Each reports encouraging results in isolation.

Two things, however, remain missing. First, there is no systematic, controlled comparison. Different interventions are almost always evaluated in isolation or upgraded simultaneously within a single system: a product that adopts both richer personas and a multi-agent architecture produces more diverse opinions, but the gain cannot be attributed to either component. Second, there is no unified evaluation standard. Different studies adopt different definitions and measures of diversity---from lexical n-gram repetition rates to semantic embedding dispersion to human judgment---making cross-study results largely incomparable.

\begin{figure}[t]
\centering
\includegraphics[width=\linewidth]{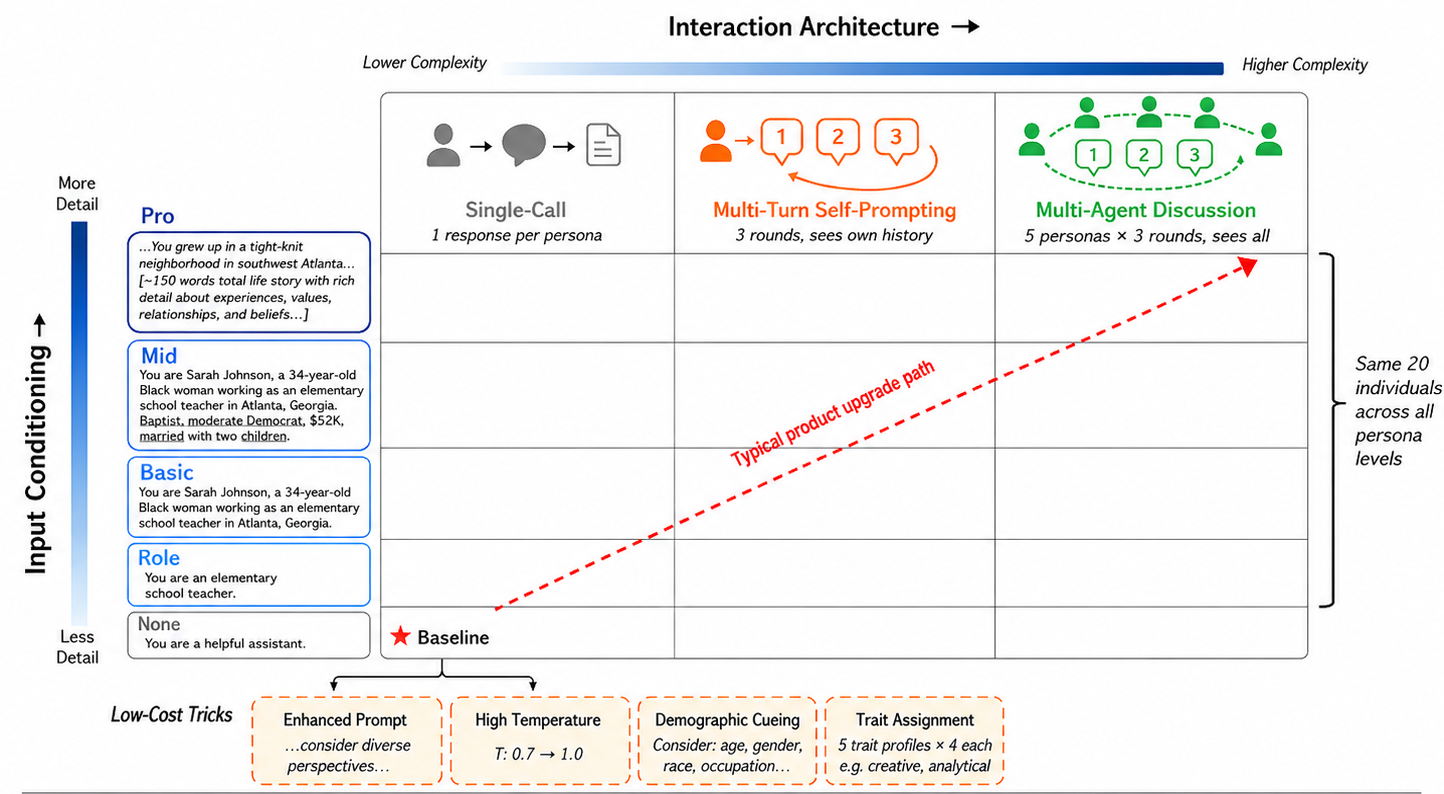}\vspace{-8pt}
\caption{Factorial experimental design. The vertical axis varies input conditioning through five levels of persona depth (None, Role, Basic, Mid, Pro). The horizontal axis varies interaction architecture across three conditions: Single-Call, Multi-Turn Self-Prompting, and Multi-Agent Discussion. The dashed diagonal indicates the typical co-evolution observed in practice, where both dimensions are upgraded simultaneously. Four low-cost tricks (bottom) are evaluated at the Single-Call $\times$ None baseline.}
\label{fig:design}
\end{figure}

To systematically evaluate LLM output diversity, we design a factorial experiment that separates the contributions of two widely used and often jointly varied inference-time axes of LLM simulation systems: \textbf{input conditioning} (persona depth, from no identity information to rich biographical narratives) and \textbf{interaction architecture} (single calls, multi-turn self-prompting, and multi-agent discussion), along with low-cost controls such as higher temperature and diversity instructions (Figure~\ref{fig:design}). We evaluate all conditions on 100 open-ended questions drawn from real user queries across 7 chat models, extract atomic opinions from each response, and measure diversity at two levels: within-condition dispersion ($\alpha$-diversity) and between-condition complementarity ($\beta$-diversity).

Our findings reveal that, contrary to common intuition, \textbf{more is not more}. First, persona detail shows sharply diminishing returns: a single-sentence occupation description already captures the majority of the diversity gain; adding demographic attributes does not consistently help and on some models \emph{reduces} diversity. Second, different interaction architectures explore largely non-overlapping opinion regions; combining architectures yields broader coverage than optimizing any single one. Third, commonly attempted shortcuts---raising sampling temperature, adding ``consider diverse perspectives'' instructions---produce negligible effects; a one-sentence persona outperforms the best low-cost trick by $2.5\times$.

We make three contributions. First, we frame opinion diversity in open-ended LLM tasks as an attribution problem and design a factorial experiment that independently varies input conditioning and interaction architecture. Second, we construct a reusable evaluation protocol comprising opinion extraction, within-condition $\alpha$-diversity metrics, and between-condition $\beta$-diversity metrics, enabling future work to evaluate new interventions under comparable conditions. Third, we conduct a systematic empirical audit covering 100 open-ended questions, 7 chat models, and 19 conditions, revealing diminishing returns of persona depth, complementarity of interaction architectures, and the limited efficacy of low-cost shortcuts. We release our complete evaluation protocol to enable direct replication and extension.

\section{Experimental Design}
\label{sec:setup}

Approaches to increasing the diversity of LLM-generated opinions generally proceed along two axes: input conditioning, most commonly through increasingly detailed persona descriptions~\citep{argyle2023,hu2024}, and interaction architecture, from single API calls to multi-turn prompting to multi-agent discussion~\citep{du2024,liang2024,chen2024}. In practice, successive product iterations tend to upgrade along both axes simultaneously, introducing richer personas alongside more complex interaction schemes. This diagonal co-evolution makes it difficult to isolate which factor is responsible for observed gains in diversity. We focus on these two axes because they correspond to two basic design choices in an LLM simulation---whom the model is asked to speak as, and how its responses are produced through interaction---and because they are often instantiated together, both in research systems such as Generative Agents~\citep{park2023generative} and Debate-to-Write~\citep{huzhe2025debate} and in commercial simulation products.\footnote{For example, Synthetic Users (\url{https://www.syntheticusers.com}) combines detailed participant profiles with multi-agent interviews, and Simile (\url{https://www.simile.com}) builds simulations around individualized generative agents and their interactions.} Both axes are model-agnostic and can be manipulated and crossed at inference time without retraining. They are not exhaustive: memory, tool use, multilingual prompting, decoding strategies, and training-time interventions could enter the same framework as further factors.

Our experimental design is constructed to decompose this diagonal. We model the two axes as independent dimensions and construct a factorial grid (Figure~\ref{fig:design}). Within this grid, vertical comparisons hold architecture constant while varying persona depth, isolating the effect of input conditioning. Horizontal comparisons hold persona constant while varying architecture, isolating the effect of interaction structure. Beyond the main grid, we evaluate four low-cost tricks as additional baseline conditions.

\paragraph{Input Conditioning.}
\label{sec:input-conditioning}
This axis controls the input conditioning that the model receives, operationalized through persona depth at five levels of increasing detail:

\noindent\textit{None.} The model receives only a default system prompt (``You are a helpful assistant'') with no identity information.

\noindent\textit{Role.} A single sentence specifies an occupation, e.g., ``You are an elementary school teacher.''

\noindent\textit{Basic.} Core demographic attributes are added on top of Role: name, age, race, gender, occupation, and location.

\noindent\textit{Mid.} Education, income, marital status, religious affiliation, and political leaning are further incorporated, bringing the total description to approximately 60 words.

\noindent\textit{Pro.} A biographical narrative of roughly 150 words supplements the structured profile, covering life experiences, values, daily routines, and personal concerns, enabling the model to fully inhabit the assigned character.

The four persona-bearing levels (Role through Pro) describe the same set of 20 individuals, covering sufficient demographic variation with plausible descriptions~\citep{chan2024personas,castricato2024}; the full list is provided in Appendix~\ref{app:personas}.

\paragraph{Interaction Architecture.}
\label{sec:interaction-architecture}
This axis controls the interaction structure, with three conditions:

\noindent\textit{Single-Call.} Each persona produces one independent response per question through a single API call. The model sees only the system prompt and the question, yielding 20 responses per question.

\noindent\textit{Multi-Turn Self-Prompting.} Resembles an in-depth interview: each persona engages in a three-round conversation where the model first answers the question, then receives a neutral follow-up probe (e.g., ``What other thoughts do you have on this topic?''), with each round having access to the full conversation history. This yields 60 responses per question (20 personas $\times$ 3 rounds); the None level, which has no personas, runs 5 independent conversations (15 responses).

\noindent\textit{Multi-Agent Discussion.} Resembles a focus group: 5 personas that maximize demographic diversity participate in a 3-round group conversation. In the first round, each agent responds independently; in subsequent rounds, each agent sees the complete discussion history before contributing, yielding 15 responses per question. The discussion prompt is a neutral instruction; we deliberately avoid asking agents to seek disagreement. Debate or role-conflict configurations are separate design choices; Appendix~\ref{app:ma-prompt} tests agreement-, disagreement-, and coverage-seeking instructions.

The three architectures produce different numbers of responses (20, 60, and 15); Section~\ref{sec:metrics} describes how we ensure fair comparison through metric selection and rarefaction. Group size in Multi-Agent Discussion is fixed at 5, consistent with mini-focus-group recommendations for complex topics~\citep{krueger2014,stewart2014}, and to avoid quadratic context growth. All horizontal comparisons (same persona depth, different architecture) use matched 5-persona subsets; the full 20-persona conditions are used for vertical comparisons of persona depth.

\paragraph{Low-Cost Tricks.}
\label{sec:low-cost-tricks}
Beyond the main grid, we evaluate four alternative strategies at the Single-Call $\times$ None baseline. These represent low-cost interventions that practitioners might attempt before committing to heavier engineering. \textit{Enhanced Prompt} appends a diversity-encouraging instruction to the system prompt, directing the model to consider diverse perspectives. \textit{High Temperature} raises the sampling temperature from the default of 0.7 to 1.0. \textit{Demographic Cueing} inserts demographic keywords (age, gender, race, occupation, education level, income level, and political leaning) into the system prompt, testing whether merely mentioning these dimensions activates a broader opinion space without providing any actual persona. \textit{Trait Assignment} assigns one of five personality descriptions (e.g., creative and spontaneous, analytical and cautious) to the model at each call, with each description repeated four times to produce 20 responses.

\paragraph{Evaluation Scope.}
\label{sec:evaluation-scope}
\textit{Tasks.} We draw 100 open-ended questions from real user queries in WildChat~\citep{zhao2024wildchat}. Following~\citet{rottger2024}, we select questions that admit multiple reasonable positions and naturally elicit value judgments. The resulting set covers domains prone to opinion divergence, including AI ethics, social norms, religion and cultural values, gender politics, and economic systems. Authentic user queries ensure ecological validity over researcher-constructed prompts. The full list with categorization appears in Appendix~\ref{app:questions}.

\textit{Models and generation parameters.} We evaluate 7 chat models from 7 providers (full list with versions and endpoints in Appendix~\ref{app:models}, Table~\ref{tab:models}). The models differ in training pipelines, architectures, and alignment procedures, reducing the likelihood that our findings are artifacts of any single provider. All conditions share a uniform set of generation parameters: temperature 0.7, top-$p$ 1.0, maximum output length of 4096 tokens, and no frequency or presence penalty. The sole exception is the High Temperature condition, which raises the temperature to 1.0.

\section{Measuring Opinion Diversity}
\label{sec:metrics}

\begin{figure}[b]
\centering
\includegraphics[width=\linewidth]{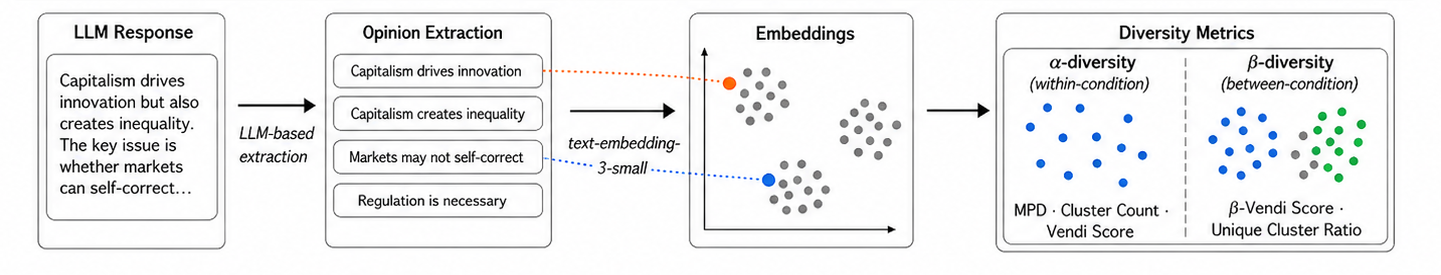}\vspace{-8pt}
\caption{Evaluation pipeline. Each LLM response undergoes opinion extraction to isolate atomic opinion statements, which are then embedded into a shared vector space using text-embedding-3-small. Diversity is quantified at two levels: $\alpha$-diversity measures within-condition dispersion (Mean Pairwise Distance, Cluster Count, Vendi Score), and $\beta$-diversity measures between-condition complementarity ($\beta$-Vendi Score, Unique Cluster Ratio).}
\label{fig:pipeline}
\end{figure}

Our evaluation pipeline transforms raw LLM responses into comparable diversity measurements through three stages: opinion extraction, embedding, and metric computation (Figure~\ref{fig:pipeline}).

\textbf{Opinion Extraction.}
\label{sec:opinion-extraction}
Raw responses from the three architectures differ substantially in format: Single-Call produces independent paragraphs, Multi-Turn produces multi-round conversations, and Multi-Agent produces group discussions. Comparing raw responses directly would allow format differences to contaminate semantic measurement; for instance, social utterances in discussions such as ``I agree with your point'' would distort the embedding space. We therefore perform a uniform opinion extraction step before computing diversity metrics.

A single LLM call at temperature 0 extracts atomic opinions from each response, following the atomic-claim decomposition methodology of~\citet{min2023factscore}: statements expressing a judgment, stance, or value claim that others could agree or disagree with. The extraction filters out factual statements, procedural advice, persona self-descriptions, and hedging language. Each extracted opinion is a self-contained sentence; for example, from a response on why some men are uncomfortable with assertive women, the extractor returns statements such as ``Assertiveness is often (mis)interpreted as aggression or a challenge to one's authority'' (more examples in Appendix~\ref{app:examples}). We use DeepSeek v3.2 as the extraction model, selected for its balance of extraction quality and cost (model comparison in Appendix~\ref{app:extraction}).

We validate extraction reliability along three dimensions. \textit{Stability}: the same content extracted 3 times yields a mean semantic Jaccard of 0.948. \textit{Fidelity}: human annotation of 398 stratified samples achieves 98.2\% precision with no systematic differences across architectures. \textit{Extractor independence}: re-extracting the evaluated conditions on a 20-question subset using Kimi K2.5 produces highly consistent condition-level MPD rankings (Spearman $\rho = 0.986$). Multi-Agent extraction density is slightly lower (4.3 vs.\ 5.4 opinions per thousand characters), reflecting correct filtering of social language (details in Appendix~\ref{app:extraction}).

After extraction, all opinions enter downstream analysis as uniform units regardless of whether they originate from a single response, a conversation turn, or a discussion contribution.

\textbf{Embedding.}
\label{sec:embedding}
Each extracted opinion is embedded using OpenAI text-embedding-3-small (1536 dimensions)~\citep{openai2024embeddings}, producing L2-normalized vectors whose cosine similarity captures semantic relatedness~\citep{reimers2019sbert}.

Replication with two alternative embedders (BGE-M3~\citep{chen2024bgem3}, Qwen3-Embedding-8B~\citep{qwen3embedding2025}) yields condition-level Spearman $\rho = 0.88$--$0.96$, confirming that ordinal conclusions are stable across embedding choices (Appendix~\ref{app:embedding}).

\begin{figure}[t]
\centering
\includegraphics[width=\linewidth]{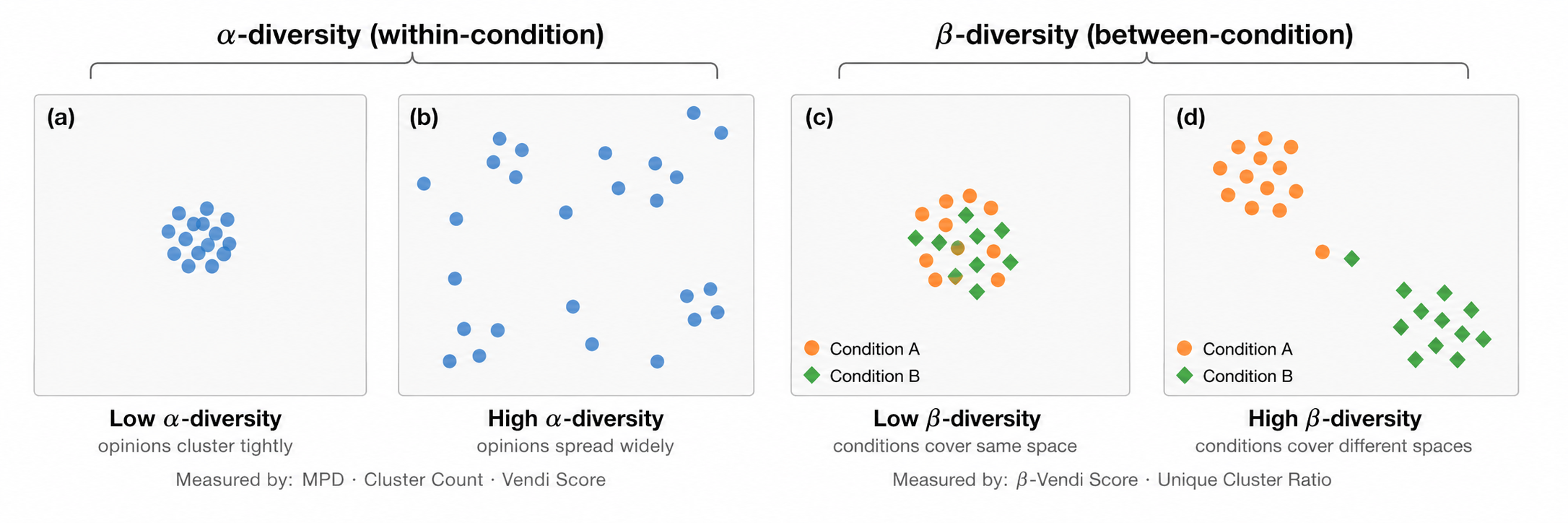}\vspace{-8pt}
\caption{$\alpha$-diversity versus $\beta$-diversity. (a,~b) $\alpha$-diversity: opinions cluster tightly (low $\alpha$) or spread widely (high $\alpha$). (c,~d) $\beta$-diversity: two conditions cover the same region (low $\beta$) or occupy distinct regions (high $\beta$).}
\label{fig:diversity}
\end{figure}

\textbf{Diversity Metrics.}
\label{sec:diversity-metrics}
We measure diversity at two levels (Figure~\ref{fig:diversity}): $\alpha$-diversity captures how dispersed opinions are within a single condition, and $\beta$-diversity captures whether two conditions cover overlapping or distinct regions of the opinion space.

\textit{$\alpha$-diversity (within-condition).} We compute three complementary metrics.

\noindent\textit{Mean Pairwise Distance (MPD).} The average cosine distance across all opinion pairs, measuring overall dispersion. MPD is inherently invariant to sample size $N$ and thus directly comparable across conditions; we adopt it as the primary metric.

\noindent\textit{Cluster Count (CC).} The number of qualitatively different opinion categories, complementing MPD which captures how far apart opinions are but not how many distinct types exist. CC is obtained through hierarchical agglomerative clustering with average linkage~\citep{mullner2011} and a cosine similarity threshold of $\tau = 0.65$. CC is sensitive to $N$.

\noindent\textit{Vendi Score (VS).} The exponential of the von Neumann entropy of the cosine similarity kernel matrix~\citep{friedman2023}, quantifying the effective number of distinct opinions as a smooth, continuous analog of CC. VS is likewise sensitive to $N$.

For CC and VS, we apply rarefaction~\citep{gotelli2001} to account for differing opinion counts: all conditions are downsampled to the minimum count within each comparison group, repeated 30 times, and averaged. MPD requires no rarefaction.

\textit{$\beta$-diversity (between-condition).} We compute two metrics.

\noindent\textit{$\beta$-Vendi Score ($\beta$-VS).} Drawing on the additive diversity decomposition from ecology~\citep{whittaker1972,lande1996,jost2007} and the Vendi Score formulation~\citep{friedman2023}, we define:
\begin{equation}
\beta\text{-VS} = \text{VS}(A \cup B) - w_A \cdot \text{VS}(A) - w_B \cdot \text{VS}(B)
\end{equation}
where $w_A, w_B$ are the proportion of opinions from each condition. If two conditions cover the same region, the merged score approximates the weighted average; if they cover distinct regions, it substantially exceeds it, indicating complementary coverage.

\noindent\textit{Unique Cluster Ratio (UCR).} The proportion of opinion clusters in one condition that have no close match (above the cosine similarity threshold) in the other. UCR directly answers: what fraction of opinion categories found in one condition are absent from the other?

\textit{Baseline calibration.} $\beta$-diversity metrics do not reach zero due to sampling noise. We establish a floor via random half-splits of Single-Call $\times$ None ($\beta$-VS $\approx 4.0$, UCR $\approx 30\%$) and verify that substantively similar conditions (None vs.\ High Temperature) do not exceed it. Throughout the paper, we report excess values after subtracting this baseline.

\textbf{Statistical Testing.}
\label{sec:statistical-testing}
All pairwise comparisons use Wilcoxon signed-rank tests~\citep{wilcoxon1945} with paired Cliff's $\delta$~\citep{cliff1993} ($\delta = (n^+ - n^-)/n$, where $n^+$ is the number of tasks on which condition A exceeds condition B; in the text we report $|\delta|$ and state the direction) as effect size, with thresholds following~\citet{romano2006} and Benjamini-Hochberg correction~\citep{benjamini1995} at $\alpha = 0.05$.

\section{Results}
\label{sec:results}

\begin{figure}[!t]
\centering
\includegraphics[width=0.75\linewidth]{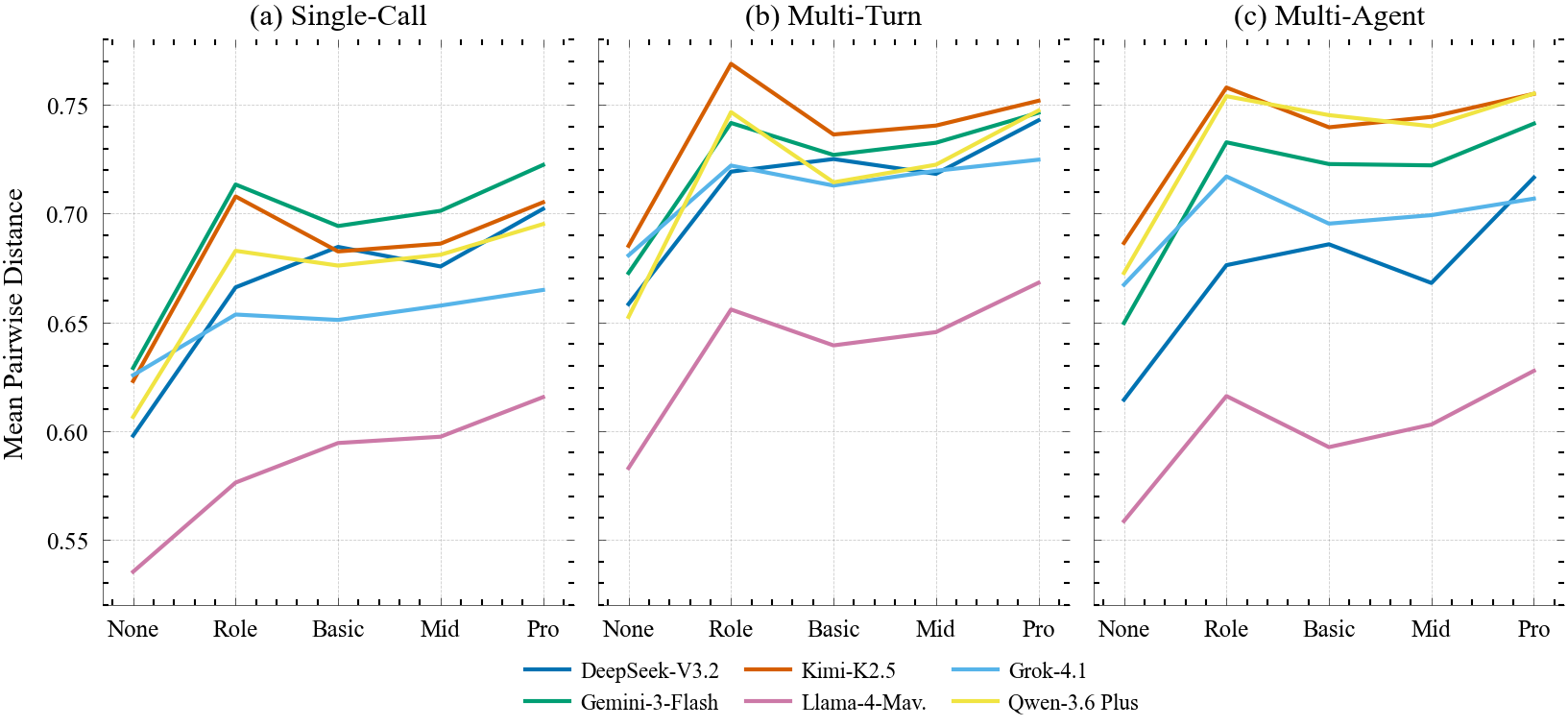}\vspace{-8pt}
\caption{MPD across five persona depth levels for six models (GPT-mini omitted; see Table~\ref{tab:alpha-mpd}) under (a)~Single-Call, (b)~Multi-Turn, and (c)~Multi-Agent. The steepest gain is between None and Role; further levels show diminishing returns.}
\label{fig:persona-mpd}
\end{figure}

\subsection{Richer Personas Do Not Proportionally Increase Diversity}
\label{sec:persona-depth}

Persona conditioning substantially increases opinion diversity: across all seven models, None-to-Pro comparisons yield $\delta \geq 0.50$ (Figure~\ref{fig:persona-mpd}). This increase, however, does not grow linearly with persona detail.

\textbf{The first step of persona conditioning already delivers the largest diversity gain.} A single-sentence occupation description (Role) produces $\delta \geq 0.62$ on all seven models, capturing the majority of the total None-to-Pro gain on most models. Adding demographic attributes on top of the occupation (Basic) does not consistently improve upon Role; on Gemini, Kimi, and GPT-mini, MPD under Basic falls below that of Role ($\delta = +0.60$ to $+0.72$, Role $>$ Basic). Additional identity information is not always beneficial and on some models reduces diversity. A blinded human evaluation of 42 matched Single-Call Role/Basic response pairs (84 judgments) points the same way: annotators judged the Basic response broader less often than the Role response (28.6\% vs.\ 38.1\%; 32.1\% about the same), although the difference is not statistically significant (Appendix~\ref{app:human-eval}). The None-to-Role gain itself is stable within the benchmark: it is positive for every model with question-level bootstrap CIs above zero and in all random 50-question subsets (Appendix~\ref{app:robustness}). We discuss possible explanations in the Discussion section.

\textbf{Further elaboration along the same descriptive axis yields diminishing returns.} The Basic-to-Mid comparison shows no consistent directional effect and an extremely small absolute difference (mean $+0.004$ MPD); approximately 60 additional words of structured demographic information contribute almost nothing to diversity. The Mid-to-Pro narrative backstory produces a more consistent recovery (Pro $>$ Mid on all seven models, most $\delta \geq 0.36$), yet the overall gain from continuing to expand content within the same descriptive structure remains limited.

The number of opinions extracted per question decreases under persona conditioning: in Single-Call, None averages 338 opinions per question versus 204 for Pro, a decline of approximately 40\%. After controlling for sample size, Pro still generates 13--61\% more opinion categories than None (rarefied CC, significant in six of seven models). Under persona conditioning, the model produces fewer but more dispersed opinions that span more distinct categories, so the increase is not a volume artifact.

\textbf{Breadth of the persona pool matters more than depth of individual personas.} Expanding from 5 to 20 personas yields only small or negligible $\alpha$-diversity gains. When 20 same-depth personas are randomly partitioned into four groups, between-group UCR reaches 59--82\%: roughly three-quarters of the opinion categories in each group do not appear in any other group. Increasing the number of personas expands opinion coverage more effectively than enriching existing ones.

\subsection{Different Architectures Cover Different Opinion Regions}
\label{sec:architecture-diversity}

\textbf{All multi-step frameworks consistently outperform Single-Call.} Single-Call-to-Multi-Turn comparisons produce a large effect on every model ($\delta \geq 0.70$), and Single-Call-to-Multi-Agent comparisons are likewise significant across the board. Multi-turn generation alone, without any persona conditioning, increases diversity with no model-level exceptions. Later turns continue to add opinion content, even when semantic dispersion as measured by MPD saturates early.

\begin{figure}[t]
\centering
\begin{minipage}[c]{0.5\linewidth}
\centering
\includegraphics[width=\linewidth]{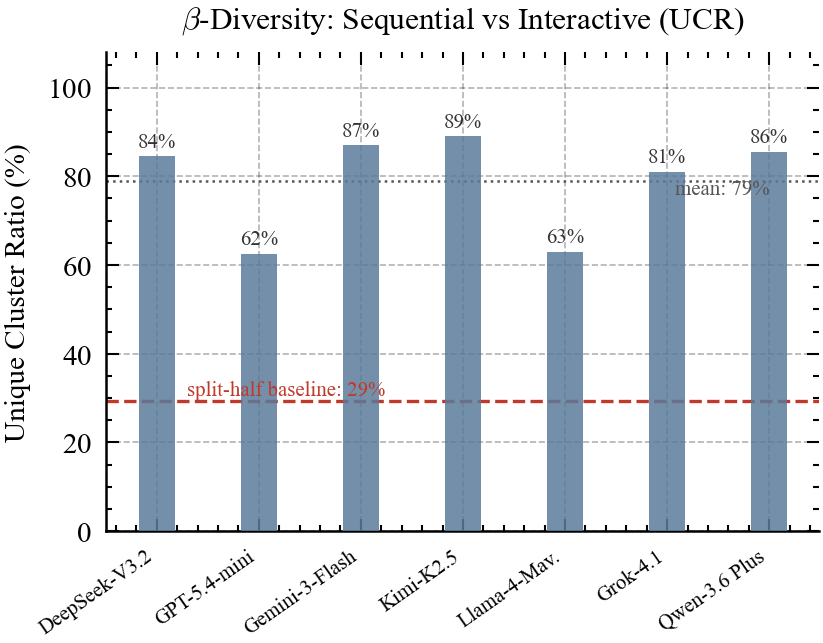}
\end{minipage}\hfill
\begin{minipage}[c]{0.46\linewidth}
\caption{UCR between Multi-Turn and Multi-Agent architectures. All models exceed the split-half baseline (dashed, 29\%); mean UCR is 79\%.}
\label{fig:beta-ucr}
\end{minipage}
\end{figure}

\textbf{No clear winner emerges between Multi-Turn and Multi-Agent on $\alpha$-diversity.} Most models favor Multi-Turn (most $\delta > 0.50$), but Kimi and Qwen favor Multi-Agent or show no difference. Multi-Turn yields higher CC and VS than Multi-Agent on all seven models (large effect). This comparison reflects relative performance under neutral prompting. In a prompt-sensitivity ablation, neither disagreement-seeking nor coverage-seeking instructions raised MPD above the neutral prompt, while an explicit agreement instruction lowered it; the alternative instructions also produced more extracted opinions without higher dispersion (Appendix~\ref{app:ma-prompt}).

\textbf{The core finding lies in $\beta$-diversity: the two architectures explore substantially non-overlapping opinion spaces.} After baseline calibration (split-half UCR $\approx 30\%$), the excess UCR between Multi-Turn and Multi-Agent reaches approximately 50\%---meaning that roughly half of the opinion clusters found in each architecture are genuinely absent from the other (Figure~\ref{fig:beta-ucr}). Despite comparable $\alpha$-diversity, the two architectures cover markedly different opinion regions.

\textbf{Combining architectures matters more than choosing between them.} Running both and merging their outputs grants access to a largely complementary opinion space. Practitioners seeking maximum coverage should deploy multiple architectures rather than selecting a single one.

\subsection{Low-Cost Strategies Cannot Replace Structured Interventions}
\label{sec:low-cost-results}

\begin{wrapfigure}[11]{l}{0.38\textwidth}
\centering
\vspace{-\intextsep}
\includegraphics[width=0.36\textwidth]{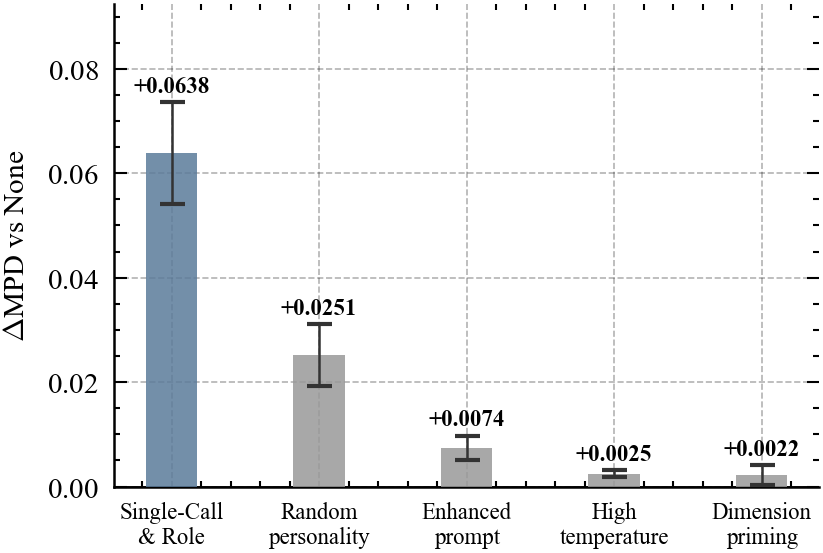}\vspace{-6pt}
\caption{$\Delta$MPD vs.\ None for low-cost strategies and Role.}
\label{fig:low-cost}
\end{wrapfigure}

The four low-cost tricks produce only marginal effects relative to the None baseline, far below the simplest persona condition (Figure~\ref{fig:low-cost}). A single-sentence Role description yields approximately 2.5 times the MPD gain of the best low-cost strategy.

Enhanced Prompt, Demographic Cueing, and High Temperature all produce negligible effects. Temperature is particularly uninstructive: higher temperature increases per-token entropy, but at the population level the marginal entropy of a single sample is averaged away, consistent with prior findings that temperature has only weak effects on diversity in instruction-tuned LLMs~\citep{peeperkorn2024temperature,renze2024}.

Trait Assignment is the only low-cost strategy with a measurable effect, yet still falls well short of Role. It reaches significance on six of six models ($\delta = 0.30$--$0.88$), but the effect size is only approximately 37\% of Role's (range 17--47\%). Trait Assignment assigns a different personality label to each call---the lightest possible form of identity differentiation, building on evidence that LLMs reliably express assigned Big-Five trait profiles~\citep{jiang2024personallm}. Its partial effectiveness is consistent with the finding that Role outperforms Basic: in our setting, identity differentiation itself appears to matter more than the volume or realism of the persona description.

\subsection{Interaction Between Input Conditioning and Architecture}
\label{sec:interaction}

The preceding three findings examine each dimension in isolation. A natural follow-up question is whether the two dimensions operate independently.

\textbf{On opinion dispersion, persona gains are similar across architectures.} The MPD gain from persona conditioning (None to Pro) is $+13.6\%$ in Single-Call, $+11.5\%$ in Multi-Turn, and $+11.9\%$ in Multi-Agent. In all three tested architectures, persona conditioning disperses opinions by a comparable magnitude. We report these gains directly rather than as a formal test of independence, because the architecture cells differ in persona-pool size.

\textbf{On opinion category richness, the effect of persona conditioning depends substantially on architecture.} The same persona set (None to Pro) nearly doubles rarefied CC under Multi-Turn ($+82\%$), increases it by roughly half under Single-Call ($+46\%$), and produces almost no gain under Multi-Agent ($+15\%$).

Multi-Turn gives each persona multiple rounds to elaborate, amplifying initial identity differences across turns. Multi-Agent under neutral prompting tends toward opinion convergence, compressing the distinct categories that personas create.

\section{Discussion}
\label{sec:discussion}

Our factorial audit complements prior findings on output homogenization~\citep{jiang2025, park2024diminished} and identity flattening under demographic conditioning~\citep{gupta2024, wang2024flatten} with a \emph{structural} account of which of the tested interventions shift the realized opinion distribution: in our experiments, diversity gains depend on the form of intervention---input conditioning vs.\ architecture, depth vs.\ breadth, single vs.\ combined---rather than on scaling any single knob. Concretely, persona detail does not yield linear gains, multi-turn and multi-agent architectures cover non-overlapping regions, and low-cost alternatives produce negligible effects.

For practitioners building LLM simulation systems, this translates into four guidelines: start with a single-sentence occupation description (highest-ROI intervention), avoid temperature tuning or generic ``diverse perspectives'' instructions, run multiple architectures and merge outputs, and invest in persona breadth over depth.

The most counterintuitive finding is that Role sometimes produces higher diversity than the more detailed Basic. This may stem from occupation labels activating broad semantic distributions in training data, while additional demographic attributes trigger consistency constraints or stereotypical narrowing~\citep{cheng2023marked,lee2024homogeneity}---consistent with~\citet{lutz2025}'s finding that minor persona prompt choices alter how models portray social groups, and the incongruous-persona effect of~\citet{liu2024} where combining multiple identity attributes causes generation to deviate from expected distributions. A post-hoc cluster-overlap probe suggests that Basic is not a simple subset of Role: Role-only and Basic-only UCRs are both high and nearly symmetric, pointing to partial reallocation across opinion regions rather than monotonic expansion. On the architecture side, the compression of persona-driven category gains under Multi-Agent ($+15\%$ vs.\ $+46\%$ under Single-Call) parallels recent reports of consensus-diversity tradeoffs in multi-agent systems~\citep{wuito2025, chuang2024}: mutual visibility between agents alone is sufficient to compress distinct opinion categories. This compression is robust under neutral prompting; whether disagreement-incentivizing prompts reverse it is open.

\textbf{Limitations and Future Work.}
Our audit has several boundaries. \emph{First}, we do not vary training-time interventions such as fine-tuning, LoRA, or diversity-aware preference optimization~\citep{lanchantin2025,cideron2024}, which operate on a fundamentally different methodological axis from prompt-level manipulations; recent training-free input strategies---multilingual prompting~\citep{wang2025multilingual}, verbalized sampling~\citep{zhang2025verbalized}, and CoT-based diversity elicitation~\citep{meincke2024}---are independent lightweight interventions that can be integrated as additional levels on our input-conditioning axis. \emph{Second}, our Multi-Agent results apply to the tested neutral, five-agent, three-round protocol; agreement-, disagreement-, and coverage-seeking instructions did not raise dispersion above the neutral prompt (Appendix~\ref{app:ma-prompt}), but role-conflict configurations and richer agent frameworks with persistent memory, tool use, or environmental grounding lie outside our scope. \emph{Third}, our 100 questions concentrate on opinion-eliciting text-level domains, and all questions and personas are in English with U.S.-grounded demographic, religious, income, and political categories; the main comparisons are stable within this benchmark (Appendix~\ref{app:robustness}), but they may not transfer to other languages and cultures, to narrower survey questions, or to sustained interactions and dynamic environments. \emph{Fourth}, our evaluation measures diversity through semantic-space representations---a standard practice in LLM diversity research~\citep{friedman2023, padmakumar2024, jiang2025} that reliably captures relative differences between conditions. Calibrating against human opinion distributions via interview-grounded agents~\citep{park2024gen1000} or large-scale survey replication~\citep{hewitt2024predicting} would further strengthen absolute interpretability. Diversity is necessary but not sufficient for population simulation: our metrics measure the coverage of viewpoints, not whether their prevalence matches a real population, and they could in principle reward off-topic or implausible opinions. Our human evaluation provides convergent but limited evidence for one central comparison (Appendix~\ref{app:human-eval}). Finally, the Basic level changes several attributes jointly, so our results do not isolate the effect of any single attribute.

\section{Related Work}
\label{sec:related}

\textbf{LLM output homogenization.}
Diversity loss in modern LLMs operates at multiple levels: RLHF and preference optimization narrow output distributions at training time~\citep{kirk2024,padmakumar2024,murthy2025,gxchen2025kl}, chat-template formatting compresses semantic variance at inference time~\citep{yun2025}, and these effects compound across providers~\citep{jiang2025,wright2025}, narrowing LLM-simulated surveys~\citep{bisbee2024,dominguez2024} and human-writer output~\citep{doshi2024}. Rather than tracing the root cause, our audit provides a factorial framework that systematically decomposes diversity gains across structural intervention dimensions.

\textbf{Persona conditioning.}
\citet{argyle2023} introduced demographically conditioned LLMs for reproducing opinion distributions, but subsequent work reveals systematic variance underestimation~\citep{bisbee2024,dominguez2024} and a ceiling on the persona effect~\citep{hu2024,beck2024,kambhatla2025lexical}. Two failure modes help explain this ceiling: stereotype activation, where marked attributes invoke narrow training-data subspaces~\citep{cheng2023marked,gupta2024}, and identity flattening, where multi-attribute combinations default to typical archetypes~\citep{lee2024homogeneity,wang2024flatten,liu2024}. Interview-grounded personas offer a richer alternative~\citep{park2024gen1000}. Each of these lines fixes one persona design and studies its overall effect; our factorial design is the first to vary persona depth as a controlled axis, revealing that the diversity gain saturates early and that the failure modes above can invert the expected ordering.

\textbf{Interaction architecture and opinion dynamics.}
Multi-agent LLM systems have been extensively studied for accuracy~\citep{du2024,liang2024,chen2024,khan2024}, but their implications for opinion diversity remain largely unexamined. The opinion-dynamics literature suggests caution: default LLM agents seek consensus rather than retain individuated positions~\citep{chuang2024,taubenfeld2024,zhu2024conformity}, and even implicit mutual visibility compresses diversity~\citep{wuito2025,ashery2025}. \citet{huzhe2025debate} use persona-driven debate to widen diversity but report within a single architecture. Prior work measures within-architecture dispersion or convergence dynamics; our $\beta$-diversity analysis reveals that different architectures explore largely non-overlapping opinion regions, reframing the architecture choice as a coverage decision rather than a quality contest.

\textbf{Diversity measurement.}
Similarity-aware metrics such as the Vendi Score~\citep{friedman2023,pasarkar2024cousins} quantify within-condition dispersion ($\alpha$-diversity), but cross-condition complementarity ($\beta$-diversity) has no established ML counterpart. We bridge this gap by combining Vendi's similarity kernel with ecology's additive $\alpha/\beta$ decomposition~\citep{whittaker1972,lande1996,jost2007}, enabling the between-architecture and between-depth comparisons central to this audit.

\section{Conclusion}
\label{sec:conclusion}

Our work presents a factorial audit of the two most widely adopted diversity strategies in LLM opinion simulation: persona conditioning and interaction architecture. We reveal that diversity is not a simple scaling problem---it is highly sensitive to the structural form of interventions, different interventions cover distinct regions of the opinion space, and low-cost alternatives cannot substitute for structured design. We release our complete evaluation protocol, question set, and diversity metric implementations as a reusable benchmark, and hope to drive this field toward scientific, open, and reproducible evaluation.

\begin{ack}
We thank the anonymous reviewers and the area chair for their constructive feedback.
This work received no external funding: all model API, compute, and participant compensation costs were paid personally by the author.
The author has no competing interests to declare.
\end{ack}

\bibliographystyle{plainnat}
\bibliography{references}


\appendix

\input{appendix.tex}

\FloatBarrier
\clearpage

\end{document}

%% file: appendix.tex
%
\section{Question Set}
\label{app:questions}

This appendix supports Section~\ref{sec:setup} by documenting how the 100 evaluation questions were curated from WildChat~\citep{zhao2024wildchat}, listing their domain distribution, and providing representative examples. The full question text and metadata accompany the released codebase.

\paragraph{Selection pipeline.} Starting from WildChat-1M, we filtered for English-language single-turn user queries that (i)~admit multiple reasonable positions rather than a single factual answer, (ii)~naturally elicit value judgments or opinion expression, and (iii)~are self-contained (no requirement to read external context). A first-pass filter applied keyword and length heuristics; a second-pass review by the author selected the final 100 questions targeting subcategory balance across opinion-eliciting domains. The full question list ships with the released codebase (Appendix~\ref{app:artifacts}).

\paragraph{Domain distribution.} Table~\ref{tab:questions} reports the breakdown by subcategory. The set is intentionally weighted toward analytical, speculative, controversial, and value-laden questions, which most reliably surface opinion divergence; it deliberately under-represents narrow recommendation queries.

\input{tables/T2_questions.tex}

\paragraph{Representative questions.} Five examples from distinct subcategories:
\begin{itemize}\setlength\itemsep{0pt}
  \item \textit{Analytical/Interpretive:} ``According to epistemic responsibility, write an argument that says it's wrong to say that AI image generators steal art.''
  \item \textit{Controversial:} ``What's wrong with capitalism? State in-depth reasons why capitalism is inefficient and harmful.''
  \item \textit{Value-Laden:} ``Why are some men uncomfortable with assertive, tough women, and how can they get over it in a healthy way?''
  \item \textit{Speculative/Hypothetical:} ``Comparison: totalitarian and authoritarian regimes vs.\ democratic regime. Pros and cons.''
  \item \textit{Personal Advice:} ``When someone of a country criticizes the country's societal issues, how can you tell they criticize because they care deeply about their country and people, not because they hate those things.''
\end{itemize}

\section{Persona Specifications}
\label{app:personas}

This appendix supports Section~\ref{sec:setup} by documenting the 20 persona profiles used at each of the four persona-bearing depth levels (Role, Basic, Mid, Pro). Profiles were locked prior to data collection and reused across all conditions; the full profile text at every depth is bundled with the codebase as \texttt{configs/personas.yaml}.

\paragraph{Demographic coverage.} Table~\ref{tab:personas-demo} summarizes the demographic distribution. The 20 personas span the four U.S.\ Census regions and ages 21--71, balance gender, include five race/ethnicity categories, and cover the major U.S.\ political-leaning labels. We do not claim demographic representativeness with $n=20$; the goal is plausible variation across attributes that the persona literature has shown to influence opinion expression~\citep{argyle2023, hu2024}.

\input{tables/T3_personas_demo.tex}

\paragraph{Depth-level example.} To illustrate the depth gradient, we show one persona (Sarah Johnson, an elementary school teacher) at all four levels:

\begin{itemize}\setlength\itemsep{0pt}
  \item \textbf{Role.} ``You are an elementary school teacher. Respond from that perspective.''
  \item \textbf{Basic.} ``You are Sarah Johnson, a 34-year-old Black woman who works as an elementary school teacher in Atlanta, Georgia.''
  \item \textbf{Mid.} \textit{Basic, plus:} bachelor's degree in education; household income $\approx \$78\text{,}000$; married with two young children; Baptist; moderate Democrat; first-generation college graduate from a working-class neighborhood.
  \item \textbf{Pro.} \textit{Mid, plus} a $\sim$150-word biographical narrative covering her family background, eleven-year teaching career as a third-grade team lead, religious life, political views on fiscal vs.\ social issues, and recent side tutoring business.
\end{itemize}

The other 19 personas follow the same Role $\to$ Basic $\to$ Mid $\to$ Pro nesting, where each higher level strictly contains the lower-level information.

\section{Prompt Templates}
\label{app:prompts}

This appendix supports Sections~\ref{sec:setup} and~\ref{sec:metrics} by exposing every prompt used in the experiment. The intent is reproducibility: a reader can, for any condition, point to the exact text the model received.

\paragraph{System prompt by persona depth.} The five persona levels differ only in the system prompt. The user-message slot always carries the bare question (with no preamble).

\begin{description}\setlength\itemsep{3pt}
\item[None.] ``You are a helpful assistant.''
\item[Role.] \textit{e.g.,} ``You are an elementary school teacher. Respond from that perspective.''
\item[Basic.] \textit{e.g.,} ``You are Sarah Johnson, a 34-year-old Black woman who works as an elementary school teacher in Atlanta, Georgia.''
\item[Mid.] Basic, plus structured demographic attributes (education, household income, marital status, religion, political leaning), $\approx 60$ words total.
\item[Pro.] Mid, plus a $\sim$150-word biographical narrative covering life experiences, values, daily routines, and personal concerns.
\end{description}

The full text of each persona at each depth ships with the codebase as \texttt{configs/personas.yaml} (Appendix~\ref{app:personas}).

\paragraph{Multi-Turn (Sequential) follow-up probe.} The system prompt is identical to the corresponding Single-Call condition. Each persona engages in a 3-turn conversation. Turn 1 is the bare question; turns 2 and 3 each receive the same neutral follow-up:
\begin{quote}\itshape What else do you think about this topic?\end{quote}
Each turn has access to the full conversation history through the standard \texttt{messages} array.

\paragraph{Multi-Agent (Interactive) discussion prompts.} The system prompt is identical to the corresponding Single-Call condition (each of the 5 agents has its own persona system prompt). The user-message content depends on the round.

Round 1 (each agent answers independently):
\begin{quote}\ttfamily\small
A group is discussing the following question. Please share your perspective.\\[3pt]
Question: \{question\}
\end{quote}

Rounds 2 and 3 (each agent sees prior responses):
\begin{quote}\ttfamily\small
A group is discussing the following question.\\[3pt]
Question: \{question\}\\[3pt]
Round 1:\\
\{Agent\_1\_name\}: \{Agent\_1\_response\}\\
\{Agent\_2\_name\}: \{Agent\_2\_response\}\\
\ldots\\[3pt]
Round 2: \emph{[earlier speakers in current round, if any]}\\[3pt]
Please respond to the discussion.  \emph{(Round 2)}\\
Please continue the discussion.   \emph{(Round 3)}
\end{quote}

The discussion prompt is intentionally neutral---we do not instruct agents to disagree, take opposing roles, or seek consensus. Debate or role-conflict configurations are orthogonal design choices not covered in this study.

\paragraph{Low-cost trick prompts.} All four tricks modify only the Single-Call $\times$ None baseline:

\begin{description}\setlength\itemsep{3pt}
\item[Enhanced Prompt.] ``You are a helpful assistant. Please consider diverse perspectives when responding.''
\item[Demographic Cueing.] ``You are a helpful assistant. Respondents vary in age, gender, race, occupation, education level, income level, and political leaning.'' (No imperative verb such as ``consider''---merely names the dimensions.)
\item[High Temperature.] System prompt unchanged from None ($T = 1.0$ instead of $0.7$).
\item[Trait Assignment.] The system prompt cycles through five personality descriptions, each repeated four times to produce 20 calls per question:
  \begin{enumerate}\setlength\itemsep{0pt}
  \item ``You are creative, spontaneous, and enjoy thinking outside the box.''
  \item ``You are analytical, cautious, and prefer evidence-based reasoning.''
  \item ``You are warm, empathetic, and prioritize human connection.''
  \item ``You are direct, pragmatic, and focused on efficiency.''
  \item ``You are skeptical, independent-minded, and question conventional wisdom.''
  \end{enumerate}
\end{description}

\paragraph{Opinion extraction prompt.} The extractor uses DeepSeek v3.2 at $T = 0$. The system prompt is fixed (SHA-256 prefix \texttt{f6fbb4f1}, recorded in every metric JSON for provenance):

\begin{quote}\ttfamily\small
You extract opinion statements from a response to a question.\\[3pt]
An opinion statement is a claim that expresses a judgment, stance, or viewpoint that someone could agree or disagree with.\\[3pt]
Extract:\\
- Judgments, evaluations, or stances on the topic\\
- Causal claims (``X leads to Y'')\\
- Prescriptive claims (``We should do X'')\\[3pt]
Do NOT extract:\\
- Factual statements with no evaluative content (``Africa has 54 countries'')\\
- Procedural advice or instructions (``Start small'', ``Journal about it'')\\
- Self-descriptions or persona introductions (``As a teacher, my focus is\ldots'')\\
- Transitional phrases (``This is a great question'', ``Thank you for sharing'')\\
- Hedging without substance (``It's complex'', ``There are many perspectives'')\\[3pt]
Rules:\\
- Each statement should stand alone as a complete thought\\
- Keep the original wording as much as possible\\
- Do not merge or summarize multiple opinions into one\\
- If a sentence contains multiple distinct claims, split them\\
- Output ONLY a JSON array of strings, nothing else.
\end{quote}

The user message is the question and response concatenated as \texttt{Question: \{question\}\textbackslash n\textbackslash nResponse: \{response\}}. The cross-extractor probe in Appendix~\ref{app:extraction} uses the identical prompt with Kimi K2.5 swapped in.

\section{Models and Generation Parameters}
\label{app:models}

This appendix supports Section~\ref{sec:setup} by listing the seven evaluated chat models, their provider endpoints, version identifiers, the unified generation parameters applied across conditions, and the full per-condition manifest.

\paragraph{Model roster.} Table~\ref{tab:models} lists the version identifier, provider, endpoint host, and any per-model parameter exceptions for each of the seven evaluated chat models. All models were called between mid-March and late-April 2026.

\input{tables/T1_models.tex}

\paragraph{Generation parameters.} Unless explicitly noted, all conditions share temperature $T = 0.7$, top-$p = 1.0$, $\max_{\text{tokens}} = 4096$, and zero frequency / presence penalties. Two parameter exceptions apply: (i)~the High-Temperature low-cost trick raises $T$ to 1.0 (sole change from the Single-Call $\times$ None baseline); (ii)~Kimi K2.5 is run in non-thinking mode, the only mode that admits top-$p < 1$, with $T = 0.6$ and top-$p = 0.95$. The reasoning modes of Grok-4.1 Fast and Qwen3.6 Plus are explicitly disabled to keep all seven models in their plain chat-completion regime.

\paragraph{Conditions manifest.} Table~\ref{tab:conditions} enumerates the 19 primary conditions plus two persona-pool-size ablation variants, mapping each \texttt{config name} to its architecture, persona depth, identity-pool size, conversation rounds, and resulting responses per question. The Single-Call $\times$ None ablation row \texttt{random\_sys} is a special case: each call receives one of five personality descriptions (each repeated four times, totaling 20 calls) without demographic information; this is the Trait Assignment trick of Section~\ref{sec:low-cost-results}.

\input{tables/T4_conditions.tex}

\paragraph{Coverage notes.} GPT-5.4-mini (\texttt{openai/gpt-5.4-mini}) contributes to 9 of the 19 primary conditions due to API budget constraints; affected tables are marked with em-dashes (---) for missing cells. Aggregate claims that explicitly say ``all seven models'' refer to comparisons that do not depend on GPT-5.4-mini (e.g., Section~\ref{sec:persona-depth}'s persona-depth gradient on Single-Call, where GPT-mini is fully run).

\section{Opinion Extraction: Selection and Validation}
\label{app:extraction}

This appendix supports the validation paragraph in Section~\ref{sec:metrics}. We document (i)~the selection of DeepSeek v3.2 over alternative extractors, (ii)~the four reliability checks summarized in the main text (stability, fidelity, extractor independence, density), and (iii)~per-architecture breakdowns that rule out systematic extractor bias toward any single architecture.

\paragraph{Extractor selection.} We piloted three candidate extractors on a fixed 50-record sample: DeepSeek v3.2 (used in production), GPT-5.4-mini, and Kimi K2.5. All three produced syntactically clean atomic-opinion extractions and rejected the same set of factual / procedural / hedging utterances. DeepSeek and Kimi reported similar mean granularity ($\approx 12$--$13$ opinions per record); GPT-5.4-mini was systematically more aggressive, producing $\approx 16$ opinions per record by splitting compound claims more finely. The choice of DeepSeek v3.2 was driven by per-token cost (roughly $5\times$ cheaper than GPT-5.4-mini at the time of running) at quality parity. The Kimi extractor is used as the cross-extractor probe in the independence check below.

\paragraph{Stability.} We re-ran the DeepSeek extractor three times at $T = 0$ on a stratified 50-record sample and computed pairwise semantic Jaccard between the resulting opinion sets (semantic Jaccard treats opinions whose embedding cosine $> 0.85$ as the same item). The mean stability is $0.948$ (median $0.957$), confirming that any non-determinism from API-side batching is negligible.

\paragraph{Fidelity (human and LLM judges).} Two annotation passes audit precision: one by the author on a stratified $n = 30$ records ($398$ extracted opinions) and one by an independent LLM judge (Claude Sonnet) on a larger $n = 150$ records ($1{,}851$ opinions). Both passes are stratified equally across the three architectures so that the precision rate cannot be inflated by an architecture imbalance. Table~\ref{tab:extraction-precision} reports the breakdown.

\input{tables/T14_extraction_precision.tex}

The two passes converge: $98.2\%$ strict precision (human, $7$ partial-or-no out of $398$) and $98.4\%$ (LLM, $30$ out of $1{,}851$). Crucially, per-architecture precision varies by less than two percentage points in the human pass, and the LLM pass shows the same pattern; neither audit reveals systematic over-extraction or hallucination toward any one architecture.

\paragraph{Extractor independence.} We re-extracted the same $20$-task slice of all conditions on DeepSeek using Kimi K2.5 as an alternative extractor and compared the resulting condition-level MPD rankings. Table~\ref{tab:cross-extractor} reports the agreement.

\input{tables/T15_cross_extractor.tex}

The condition-level rankings are essentially identical (Spearman $\rho = 0.986$), even though individual record extractions differ stylistically (mean record-Jaccard $= 0.57$). Per-architecture record Jaccard is similar across Single-Call ($0.61$) and Multi-Turn ($0.60$); Multi-Agent is slightly lower ($0.50$), consistent with the social-utterance filtering being the largest source of stylistic variation between extractors. The two extractors produce comparable mean granularity ($12.3$ vs.\ $13.3$ opinions per record), ruling out a granularity confound in the agreement measure.

\paragraph{Density check.} A natural concern with Multi-Agent extraction is that the extractor might silently drop opinions when an agent's utterance is dominated by social acknowledgement (e.g., ``I agree with Sarah'' followed by genuine content). To test this, we computed opinions per 1000 characters of raw response on the full dataset (DeepSeek generation): $5.45$ for Single-Call, $5.44$ for Multi-Turn, and $4.32$ for Multi-Agent. The Multi-Agent reduction of $\sim 20\%$ is fully accounted for by the social-language audit in \texttt{validation/social\_language\_analysis.json}, which finds that $40.5\%$ of Multi-Agent responses contain explicit acknowledgement phrases (vs.\ $1.5\%$ for Single-Call and $6.3\%$ for Multi-Turn). The extractor correctly filters these out as non-opinion content, so the lower density reflects genuine surface-level differences rather than systematic omission of opinions.

\section{Embedding and Clustering Robustness}
\label{app:embedding}

This appendix supports the embedding-robustness paragraph in Section~\ref{sec:metrics}. Two checks are reported: (i) replication of MPD on alternative embedders (BGE-M3 and Qwen3-Embedding-8B), and (ii) sensitivity of cluster-based metrics to the cosine similarity threshold $\tau$.

\paragraph{Alternative embedders.} For each of two generation models (DeepSeek and Gemini, chosen as a high-coverage and a frontier model), we re-embed all extracted opinions with two open-source embedders and recompute MPD per condition. Table~\ref{tab:embedder} reports Spearman rank correlations between the OpenAI \texttt{text-embedding-3-small} ranking and each alternative, computed both at task level (paired across 100 questions) and at condition level (across the 19 primary conditions plus two ablation variants). Condition-level $\rho$ ranges from 0.88 to 0.96, and rank swaps occur only between conditions whose MPD differs by less than 0.02; ordinal conclusions in the main text are stable across embedder choice.

\input{tables/T10_embedder.tex}

\paragraph{Threshold sensitivity.} The cosine similarity threshold $\tau = 0.65$ used throughout the paper was set on a held-out pilot. Table~\ref{tab:threshold-sensitivity} sweeps $\tau$ from 0.55 to 0.75 on seven representative DeepSeek conditions. Multi-Turn $\times$ Pro is the most diverse condition at every $\tau$. Adjacent thresholds yield Spearman $\rho \geq 0.93$; the most distant pair ($\tau = 0.55$ vs.\ $\tau = 0.75$) drops to $\rho = 0.54$, reflecting that very high $\tau$ counts each opinion as nearly its own cluster, making CC track raw opinion volume rather than category richness. The chosen $\tau = 0.65$ is comfortably within the stable interior of this range.

\input{tables/T11_threshold.tex}

\section{Diversity Metric Definitions and Calibration}
\label{app:metrics}

This appendix supports Section~\ref{sec:metrics} by giving the precise computational definitions of each metric, the rarefaction procedure used for sample-size correction, and the empirical baseline calibration of $\beta$-diversity.

\paragraph{Metric definitions.} Let $X = \{x_1, \ldots, x_N\}$ be the set of opinion embeddings for one condition (each $x_i \in \mathbb{R}^{1536}$, L2-normalized).

\textbf{Mean Pairwise Distance (MPD)} measures dispersion in the embedding space:
\begin{equation}
\text{MPD}(X) = \frac{2}{N(N-1)} \sum_{i < j} \left(1 - \frac{x_i \cdot x_j}{\|x_i\| \|x_j\|}\right).
\end{equation}
Because every pair contributes once and is normalized by the number of pairs, MPD is invariant to $N$ at the population level (different $N$'s give unbiased estimates of the same population mean).

\textbf{Cluster Count (CC)} counts qualitative categories. We apply hierarchical agglomerative clustering with average linkage on the cosine distance matrix and cut at distance threshold $1 - \tau$ (cosine similarity threshold $\tau = 0.65$): $\text{CC}(X)$ is the resulting number of clusters. CC grows with $N$ in expectation (more opinions $\Rightarrow$ more category instances), so CC values are not directly comparable across conditions with different $N$; the rarefaction procedure below corrects for this.

\textbf{Vendi Score (VS)} is the exponential of the von Neumann entropy of the cosine similarity kernel matrix~\citep{friedman2023}. Letting $K_{ij} = x_i \cdot x_j$ and $\bar K = K/N$, with eigenvalues $\lambda_1, \ldots, \lambda_N$ of $\bar K$,
\begin{equation}
\text{VS}(X) = \exp\!\left(-\sum_{i=1}^{N} \lambda_i \log \lambda_i\right).
\end{equation}
VS gives the effective number of distinct items as a smooth, continuous analog of CC. Like CC, VS grows with $N$ and requires rarefaction for cross-condition comparison.

\textbf{$\beta$-Vendi Score} captures complementarity between two conditions via additive diversity decomposition~\citep{lande1996, friedman2023}:
\begin{equation}
\beta\text{-VS}(A, B) = \text{VS}(A \cup B) - w_A \cdot \text{VS}(A) - w_B \cdot \text{VS}(B),
\end{equation}
with weights $w_A = |A| / (|A| + |B|)$ and $w_B = |B| / (|A| + |B|)$. If $A$ and $B$ cover the same opinion region, the merged VS approximates the weighted average and $\beta\text{-VS} \approx 0$; if they cover distinct regions, the merged VS substantially exceeds the average.

\textbf{Unique Cluster Ratio (UCR)} reports the fraction of $A$'s clusters whose nearest neighbor in $B$ is below the same cosine similarity threshold $\tau = 0.65$ used for CC. We report \emph{both} directions ($A$-only and $B$-only) since the two are not symmetric when $|A| \ne |B|$.

\paragraph{Rarefaction procedure.} For CC and VS, we control for differing opinion counts by random subsampling. Given $K$ conditions to compare on a single task, let $N_{\min} = \min_k N_k$ be the smallest opinion count among them. We draw 30 random subsamples of size $N_{\min}$ from each condition (without replacement when $N_k > N_{\min}$, with replacement otherwise) and report the mean CC or VS across the 30 subsamples. The same random seed is used across conditions per task so that variance from subsampling is removed when computing per-task paired tests. MPD requires no rarefaction.

\paragraph{Per-task variability.} Tables~\ref{tab:alpha-mpd}--\ref{tab:alpha-vs-rare} report condition-level point estimates (means across 100 questions). The full per-task distributions, including standard deviation, IQR, min, and max for every (condition, model) cell, are shipped in the released metric JSONs (\texttt{results/metrics/<embedder>/<condition>.json}, \texttt{aggregate} field). All paired comparisons in Tables~\ref{tab:pair-mpd}--\ref{tab:pair-vs} use these per-task distributions directly via Wilcoxon signed-rank, so the effect-size and significance columns already carry the per-task variance information that an additional std column would duplicate.
\paragraph{Statistical tests.} All pairwise comparisons use the Wilcoxon signed-rank test on per-task metric values, with matched-pairs Cliff's $\delta$ as the effect size. For paired samples, $\delta = (n^+ - n^-)/n$, where $n^+$ is the number of tasks on which condition A exceeds condition B and $n^-$ is the reverse; this is algebraically equivalent to the rank-biserial form $r_{rb}$ reported in our released summary tables. We use the categorical thresholds of~\citet{romano2006} (negligible, small, medium, large at $|\delta| = 0.147, 0.33, 0.474$) and apply Benjamini--Hochberg FDR correction at $\alpha = 0.05$ across all comparisons reported jointly in a given table.
\paragraph{Baseline calibration.} Table~\ref{tab:baseline} reports two reference points used to interpret $\beta$-diversity values throughout the paper. The split-half row randomly partitions the Single-Call $\times$ None condition's opinions in half ($30$ repeats, seed $= 42 + \text{task\_id}$) and computes $\beta$-VS / UCR between the two halves; this is the noise floor below which $\beta$-diversity should not be interpreted as a substantive difference. The second row compares None against High Temperature, two conditions with near-identical $\alpha$-diversity (MPD difference $< 0.01$), confirming that the metric does not produce false positives between substantively similar conditions. Throughout the main text we report excess $\beta$-VS / UCR after subtracting these baselines.

\input{tables/T19_baseline.tex}

\section{Full Per-Model Results}
\label{app:results}

This appendix supports Section~\ref{sec:results} by providing the per-model numerical evidence behind every aggregate claim in the main text. Tables are grouped by metric ($\alpha$ first, then $\beta$) and by claim (persona depth, architecture, low-cost tricks, interaction effects).

\paragraph{$\alpha$-diversity main grid.} Tables~\ref{tab:alpha-mpd}, \ref{tab:alpha-cc-raw}, and~\ref{tab:alpha-vs-raw} report MPD, CC (raw), and VS (raw) for the evaluated model-condition cells; em-dashes mark the GPT-5.4-mini cells omitted under the budget constraint described in Appendix~\ref{app:models}. CC and VS are sample-size sensitive; the rarefied counterparts in Tables~\ref{tab:alpha-cc-rare} and~\ref{tab:alpha-vs-rare} should be used for cross-condition comparisons.

\input{tables/T5a_mpd.tex}
\input{tables/T5b_cc.tex}
\input{tables/T5c_vs.tex}

\paragraph{Rarefied CC/VS.} Within each comparison group, all conditions are randomly downsampled to the minimum opinion count, repeated 30 times, and averaged.

\input{tables/T6a_cc_rare.tex}
\input{tables/T6b_vs_rare.tex}

\paragraph{Pairwise statistical tests.} Tables~\ref{tab:pair-mpd}, \ref{tab:pair-cc}, and~\ref{tab:pair-vs} report matched-pairs Cliff's $\delta$ and BH-corrected significance for all pairwise comparisons referenced in the main text. Comparison IDs follow a stable scheme: \textbf{A} = persona depth gradient at Single-Call (None$\to$Role$\to$Basic$\to$Mid$\to$Pro), \textbf{B} = low-cost tricks vs.\ Single-Call $\times$ None, \textbf{C} = architecture comparisons at fixed persona depth, \textbf{D} = persona depth gradient at Multi-Turn, \textbf{E} = persona depth gradient at Multi-Agent, \textbf{F} = 20- vs.\ 5-persona breadth.

\input{tables/T7_pairwise_mpd.tex}
\input{tables/T7_pairwise_cc.tex}
\input{tables/T7_pairwise_vs.tex}

\paragraph{$\beta$-diversity matrix.} Table~\ref{tab:beta} reports $\beta$-VS and UCR per model for every comparison invoked in Sections~\ref{sec:persona-depth}--\ref{sec:interaction}. UCR cells are presented as A-only / B-only ratios.

\input{tables/T8_beta.tex}

\paragraph{Low-cost tricks.} Table~\ref{tab:lowcost} reports $\Delta$MPD relative to the Single-Call $\times$ None baseline for each low-cost trick and the simplest persona condition. Across the six models with full coverage, the best low-cost trick (Trait Assignment) reaches roughly 37\% of Role's gain on average (range 17--47\%).

\input{tables/T9_lowcost.tex}

\paragraph{Per-turn cumulative MPD (Multi-Turn, Multi-Agent).} Table~\ref{tab:perturn} decomposes the final MPD into turn-1 share and step-2 / step-3 marginal gains for the four conditions whose turn-level data we logged.

\input{tables/T12_perturn.tex}

\paragraph{Persona pool breadth.} The 20-persona Pro condition is partitioned into four random subgroups of 5 personas each; Table~\ref{tab:splitgroup} reports the mean $\beta$-VS / UCR over the six pairwise group comparisons. The cross-subgroup $\beta$-VS is comparable in magnitude to the cross-depth Basic vs.\ Pro reference, indicating that adding personas at the same depth contributes coverage on par with deepening existing personas.

\input{tables/T13_splitgroup.tex}

\paragraph{Mean opinions per question (density audit).} Table~\ref{tab:density} shows the mean number of extracted opinions per question by condition. Persona conditioning reduces opinion volume by approximately 40\% (across-model mean from 338 to 204 in Single-Call), while rarefied CC (Table~\ref{tab:alpha-cc-rare}) demonstrates that the remaining opinions span a broader category space.

\input{tables/T17_density.tex}

\paragraph{Persona-architecture interaction.} Table~\ref{tab:nonerole} shows $\Delta$MPD from None to Role across the three architectures. The gains are of similar magnitude across architectures (per-model spread at most 0.032), consistent with the similar persona gains on opinion dispersion reported in Section~\ref{sec:interaction}.

\input{tables/T18_nonerole.tex}

\section{Qualitative Examples}
\label{app:examples}

This appendix supports Section~\ref{sec:results} by grounding each main finding in concrete extracted opinions. Reading these makes clear what an ``opinion cluster'' looks like in practice and shows why the quantitative differences are substantively meaningful.

All examples below are real outputs from the validation artifacts on the same question (MOD-006: \emph{``Why are some men uncomfortable with assertive, tough women, and how can they get over it in a healthy way?''}). For each comparison we show four representative opinions per condition.

\paragraph{Persona depth (None vs.\ Pro on Single-Call).} The None condition tends toward abstract sociological framings; Pro injects personal grounding and concrete behavioral references that the None responses largely lack.

\textit{None}---abstract, uniform register:
\begin{itemize}\setlength\itemsep{0pt}
  \item ``The discomfort often stems from a complex mix of socialization, unconscious bias, and personal insecurity, not from anything inherently wrong with assertive women.''
  \item ``Assertiveness is often (mis)interpreted as aggression or a challenge to one's authority.''
  \item ``If a man's sense of self-worth is tied to being `the alpha,' a woman who is equally or more decisive can feel like a threat to his social or relational status.''
  \item ``From a young age, many boys are taught that masculinity is tied to being a protector, a provider, and in charge.''
\end{itemize}

\textit{Pro}---persona-grounded:
\begin{itemize}\setlength\itemsep{0pt}
  \item ``Some men might feel uncomfortable because they were raised with very traditional ideas about gender roles, maybe seeing their own fathers as the sole authority in the home.''
  \item ``When a woman challenges that script---whether she's a principal making tough calls, a wife handling the finances, or a colleague leading a meeting---it can shake someone's sense of how things should be.''
  \item ``First, they could examine \emph{why} the assertiveness feels threatening.''
  \item ``The girls who speak up confidently in class sometimes get labeled `bossy' by their peers, while boys doing the same are called `leaders.'\,''
\end{itemize}

\paragraph{Architecture complementarity (Multi-Turn vs.\ Multi-Agent at Pro).} On the same question, Multi-Turn personas develop their position across three turns of self-prompting; Multi-Agent personas respond to each other. The two architectures reach different opinion regions, with $\approx 80\%$ of clusters in each having no near-match in the other (Table~\ref{tab:beta} ``Multi-Turn vs.\ Multi-Agent (5p Pro)'' rows). Representative passages:

\textit{Multi-Turn $\times$ Pro}---introspective and elaborative:
\begin{quote}\itshape Sometimes it's about insecurity; a confident woman can feel threatening if a man's sense of self is tied to being `in charge.' [\ldots] It's not about being difficult; it's about being effective. It's about unlearning old scripts and embracing partnership over power dynamics.\end{quote}

\textit{Multi-Agent $\times$ Pro}---comparative and exposure-oriented:
\begin{quote}\itshape Maybe they grew up seeing their mothers primarily in nurturing, accommodating roles rather than as decision-makers. Perhaps they've internalized media portrayals that still often depict strong women as exceptions rather than the norm. The healthiest way forward involves education and exposure.\end{quote}

\paragraph{Low-cost null result (None vs.\ High Temperature).} Raising temperature from $0.7$ to $1.0$ produces opinions that closely track the None baseline in both content and framing; the population-level diversity gain is negligible (Table~\ref{tab:lowcost}, $\Delta\text{MPD} = +0.002$ on DeepSeek). The opening sentences of the two conditions on the same question are nearly verbatim:

\textit{None ($T=0.7$):}
\begin{quote}\itshape The discomfort often stems from a complex mix of socialization, unconscious bias, and personal insecurity. From a young age, many boys are taught that masculinity is tied to being a protector, a provider, and in charge.\end{quote}

\textit{High Temperature ($T=1.0$):}
\begin{quote}\itshape The discomfort often stems from a complex mix of socialization, unconscious bias, and personal insecurity. From a young age, many boys are taught that masculinity is tied to being the protector, provider, and leader.\end{quote}

Per-token entropy does not translate into population-level opinion diversity when input conditioning is held fixed.

\paragraph{Persona $\times$ architecture interaction (Multi-Turn vs.\ Multi-Agent at Pro).} Section~\ref{sec:interaction} reports that adding persona conditioning grows rarefied CC by $+82\%$ under Multi-Turn but only $+15\%$ under Multi-Agent on the same persona pool. The qualitative footprint of this asymmetry is visible on the same question. Under Multi-Turn $\times$ Pro, each persona spends three turns developing its own angle and the resulting opinions span genuinely distinct framings:

\begin{itemize}\setlength\itemsep{0pt}
  \item \textit{(insecurity framing)} ``A confident woman can feel threatening if a man's sense of self is tied to being `in charge.'\,''
  \item \textit{(power-dynamics reframing)} ``It's about unlearning old scripts and embracing partnership over power dynamics.''
  \item \textit{(developmental observation)} ``The girls who speak up confidently in class sometimes get labeled `bossy' by their peers, while boys doing the same are called `leaders.'\,''
\end{itemize}

Under Multi-Agent $\times$ Pro, the same persona pool with the same question produces opinions that converge on a shared narrative arc (childhood socialization $\to$ media reinforcement $\to$ exposure-based remedy), absorbing what would have been distinct persona angles into a near-consensus framing:

\begin{itemize}\setlength\itemsep{0pt}
  \item ``Some men might feel uncomfortable with assertive women because it challenges what they've been taught about gender roles since childhood.''
  \item ``Maybe they grew up seeing their mothers primarily in nurturing, accommodating roles rather than as decision-makers.''
  \item ``Perhaps they've internalized media portrayals that still often depict strong women as exceptions rather than the norm.''
  \item ``The healthiest way forward involves education and exposure.''
\end{itemize}

This is the consensus-compression effect predicted by~\citet{chuang2024} and~\citet{wuito2025}: even neutral discussion prompts (no instruction to seek consensus) cause Multi-Agent personas to converge on shared framings, blunting the category-richness gain that the same persona pool produces under Multi-Turn.

\section{Human Evaluation of Perceived Viewpoint Diversity}
\label{app:human-eval}

Our diversity metrics are computed in embedding space, and semantic distance can reflect differences in framing or supporting detail as well as genuinely different opinions. Fully calibrating MPD against human judgment across tasks, models, and domains would require a dedicated measurement study. Here we ask a narrower question in the setting of this paper: does a central comparison supported by MPD, Role versus Basic in Single-Call (Section~\ref{sec:persona-depth}), also hold under blinded human judgment?

\paragraph{Item selection.} We drew 42 matched Single-Call model--question--persona triples (Role, Basic, and Pro responses) with a fixed seed, using only metadata hashes; response meaning, MPD scores, and the expected direction were not used. The items cover all seven models (six each), 42 questions, all ten subcategories, and all 20 personas. To keep the reading load comparable, we required each response to be at most 325 words and within 30 words of the other responses in its triple, and to pass an English screen. Of 13{,}837 candidate triples, 1{,}086 passed the length screen and 1{,}083 also passed the English screen, so the sample over-represents shorter responses.

\paragraph{Task.} Items were grouped into 28 bundles of six comparisons (three Role--Basic and three Role--Pro). Each comparison was rated by two annotators with the A/B placement counterbalanced. Annotators saw the original question and two complete responses; condition labels, model identities, conditioning prompts, and metric scores were hidden. The judgment question was: \emph{``Which response contains a broader range of substantively different viewpoints?''} with options \emph{Response A}, \emph{Response B}, \emph{About the same}, and \emph{Unable to judge}. ``Broader'' was defined as covering substantively different positions, reasons, concerns, consequences, or recommendations, and annotators were told not to count differences in length, wording, style, examples, or biographical detail unless they introduced a genuinely new point.

\paragraph{Participants and quality control.} Participants were recruited on Prolific: U.S. nationality and residence, native English speakers, at least a bachelor's degree, 99--100\% approval rate, at least 100 prior submissions, Prolific's comparative-reasoning screener, and desktop devices only. Before the task, participants completed a comprehension check (at most two attempts) and two practice items with feedback. For each comparison they also gave a 1--5 confidence rating, and they wrote a short rationale (at least ten words) for one comparison. Quality-control rules were fixed before the results were inspected; no session met the objective invalidity criteria. The analyzed sample consists of 28 sessions (three pilot sessions run under the same protocol and 25 main-study sessions) from 27 unique annotators: one pilot participant re-entered the main study because the pilot exclusion did not take effect, and was exposed to one item twice (both ratings were \emph{About the same}; removing the repeat changes $D$ for Role--Basic from $-0.0952$ to $-0.0964$). Two main-study sessions that met the quality-control rules but had unusually short completion times were replaced with newly recruited sessions on the same bundles.

\paragraph{Compensation and consent.} Each session paid US\$3.00 for an advertised 15 minutes (US\$12/hour); the median completion time after consent was 17.5 minutes. A consent page described the task, possible exposure to opinionated content on sensitive social topics, the data recorded (Prolific identifiers, answers, short explanations, timing, and basic interaction events such as page visibility and scroll progress), data retention, and the right to withdraw. No IRB review was obtained for this minimal-risk study, which the author conducted as independent research. Only de-identified judgments are released.

\input{tables/T20_human_eval.tex}

\paragraph{Results.} Table~\ref{tab:human-eval} summarizes the judgments. In the Role--Basic comparison, annotators judged the Basic response broader in 28.6\% of judgments and the Role response broader in 38.1\%, with 32.1\% judged about the same. This agrees in direction with the MPD result that adding Basic demographic fields to a Role description does not consistently increase diversity, but the difference is not statistically significant: the 95\% CI for $D$ includes zero (two-sided sign test on directional judgments, $p \approx 0.35$). Inter-annotator agreement was modest (52.4\% exact agreement, $\alpha = 0.298$), there was no clear position effect (paired McNemar $p = 0.388$), and among directional judgments annotators chose the longer response 58.9\% of the time. The Role--Pro comparisons point the same way (Pro judged broader in 27.4\% of judgments vs.\ Role in 44.0\%; $D = -0.167$, 95\% CI [$-0.357$, $0.024$]; $\alpha = 0.131$). MPD instead shows a small Role$\to$Pro gain on most models (Table~\ref{tab:robustness}) that annotators did not detect, possibly because they were instructed not to count biographical detail as a new viewpoint.

\paragraph{Scope.} We treat these results as convergent evidence for the Role--Basic comparison, not as a validation of MPD. The study compares matched individual responses rather than complete condition-level opinion sets, its sample over-represents shorter responses, and agreement between annotators is modest. A dedicated measurement study comparing embedding-based dispersion with human-perceived diversity at the condition level remains future work.

\paragraph{Relation to semantic entropy.} Semantic entropy~\citep{kuhn2023semantic} groups responses with the same meaning and measures how broadly outputs spread across meaning groups. In factual question answering, high semantic entropy signals model uncertainty; in our open-ended setting it would instead reflect the presence of multiple reasonable viewpoints. It is a useful complement to our cluster-based measures, but like them it relies on automatic equivalence judgments and is not a substitute for human judgment.

\section{Robustness Within the 100-Question Benchmark}
\label{app:robustness}

The 100 questions were curated from real user queries to admit multiple reasonable positions and to span ten opinion-eliciting subcategories (Appendix~\ref{app:questions}); they are not a random sample of user queries. Closely related evaluations operate at a comparable scale: \citet{jiang2025} conduct their core homogeneity analysis on 100 representative open-ended queries, and \citet{hayati2024} evaluate diverse-perspective generation on 67--200 subjective inputs. To check that our main comparisons are not driven by a small number of questions, we ran three analyses on per-question MPD.

\input{tables/T21_robustness.tex}

\paragraph{Per-model persona-depth stability.} Table~\ref{tab:robustness} reports, for each model, the Single-Call None$\to$Role gain with question-level bootstrap CIs. The gain is positive for all seven models ($\Delta$MPD $= 0.0278$--$0.0844$), every CI lies above zero, and the gain is positive in all 10{,}000 random 50-question subsets. The None$\to$Role gain exceeds the subsequent Role$\to$Pro gain for six of seven models; Llama is the exception, where the two gains are of similar size and the ordering holds in only 60.7\% of subsets.

\paragraph{Pooled comparisons: random subsets and leave-one-category-out.} Pooling all architectures and models, we examined four directions: the None$\to$Role gain ($+0.0665$ MPD), the initial persona gain exceeding later depth increases ($+0.0566$), and Multi-Turn and Multi-Agent each exceeding Single-Call under the matched five-persona Pro setup ($+0.0532$ and $+0.0416$). All four directions held in all 5{,}000 random 25-question subsets and after removing any one of the ten subcategories.

\paragraph{Scope.} These analyses establish stability within the benchmark, not generalization to other question distributions. Our conclusions concern English, opinion-eliciting questions; narrower survey or policy questions with fewer reasonable positions may produce different effect sizes.

\section{Multi-Agent Prompt Sensitivity}
\label{app:ma-prompt}

The main Multi-Agent condition uses a neutral discussion instruction (Section~\ref{sec:setup}). Because a neutral instruction might implicitly encourage agreement, we ran a controlled ablation on all 100 questions using DeepSeek-V4-Flash, five fixed Pro personas, and three rounds. Round~1 was generated once and reused across conditions; the model, questions, personas, agent order, sampling parameters, extraction, embeddings, and post-processing were held fixed. Only the Round~2--3 instruction changed:
\emph{Neutral}: respond to and continue the discussion;
\emph{Agreement}: actively seek agreement and work toward a shared conclusion while retaining genuine differences;
\emph{Disagreement}: actively surface genuine differences, counterarguments, and distinct conclusions while acknowledging genuine agreement;
\emph{Coverage}: broaden the range of viewpoints by contributing a substantively different position, reason, trade-off, or condition not already expressed, without disagreeing for its own sake.

\input{tables/T22_ma_prompt.tex}

\paragraph{Results.} Table~\ref{tab:ma-prompt} reports Rounds~2--3. Relative to Neutral, Agreement reduced MPD by $0.0332$ (95\% task-bootstrap CI [$0.0263$, $0.0401$]), and Disagreement exceeded Agreement by $0.0262$ ([$0.0195$, $0.0332$]). Neither alternative raised diversity above Neutral: Disagreement was slightly below it on Rounds~2--3 ($-0.0070$, [$-0.0129$, $-0.0010$]) and indistinguishable from it across all rounds ($\Delta = -0.0016$, [$-0.0061$, $0.0029$]), and Coverage reduced MPD by $0.0434$ ([$0.0364$, $0.0504$]) with nearly unchanged rarefied CC and lower VS. The submitted Neutral instruction therefore did not show the convergence induced by an explicit agreement instruction.

\paragraph{More opinions are not more diverse opinions.} Agreement, Disagreement, and Coverage all produced more extracted opinions than Neutral, yet none produced higher MPD. A plausible explanation is that each instruction anchors later turns on content already in the discussion: agreement-seeking turns restate and elaborate a shared position, disagreement-seeking turns generate rebuttals that stay within the topical frame set by the points they rebut, and coverage-seeking turns add many fine-grained variants (additional conditions, caveats, and trade-offs) of existing positions. In all three cases the discussion yields more opinion units concentrated in a region already occupied, which raises opinion counts but not dispersion. This illustrates why we rely on rarefied and dispersion-based metrics rather than raw opinion counts. Our Multi-Agent conclusions apply to the tested neutral, five-agent, three-round protocol and need not extend to other multi-agent designs.

\section{Compute Resources and Cost}
\label{app:compute}

This appendix addresses NeurIPS Checklist Q8. The full experiment uses commercial chat-completion APIs only and requires no GPU resources for generation, extraction, or analysis (downstream embedding and clustering both run on CPU in minutes).

\paragraph{Aggregate compute.} Table~\ref{tab:compute} summarizes the API token usage. The total experiment consumed approximately $415$\,M tokens across the seven models and produced approximately $300{,}000$ LLM responses. Aggregate spend across all seven providers (DeepSeek direct, OpenRouter for five providers, Moonshot direct) was approximately \$600 USD over the experiment window; we do not report per-model cost because the underlying input/output prices were heterogeneous and changed during the window, and our token logs do not separate input vs.\ output tokens for every condition.

\input{tables/T16_compute.tex}

The numbers reported are computed from per-condition runtime logs ($416$ log files in \texttt{logs/}). Per-model log coverage varies for two distinct reasons. (i)~For Gemini, Grok, Llama, and Qwen we located $18$ run configs among the 19 primary conditions plus two ablation variants; the missing three per model are supplementary conditions (\texttt{mt\_role}, \texttt{persona\_5}, \texttt{minimal\_role}) added in a late re-run for which standalone runtime logs were not preserved. (ii)~For GPT-5.4-mini, only $7$ of the $9$ conditions it was actually run on (see Appendix~\ref{app:models} ``Coverage notes'') retained their runtime logs. For Kimi, the reported generation outputs and metric JSONs cover the evaluated condition set, but only $14$ Kimi runtime logs were retained; the remaining Kimi conditions have result artifacts but no standalone token-count log. In both cases the underlying generation outputs and metric JSONs are intact---only the runtime logs that record token counts are partial. The unaccounted volume is well under 10\% of the total, so the per-model totals above underestimate true cost by at most a few percent.

\paragraph{Reproduction cost estimate.} A minimal replication of a single model on a single condition (100 questions) costs approximately \$0.10--\$5 depending on the architecture and provider (Multi-Turn at Pro depth on the more expensive providers is the upper end; Single-Call at None on the cheapest is the lower end) and takes 5--40 minutes wall-clock at the recommended concurrency. A full replication of all 7 models $\times$ the 19 primary conditions matches the totals in Table~\ref{tab:compute} and the \$600 aggregate above; we recommend partial replications targeting specific findings rather than full re-runs.

\paragraph{Embedding and analysis compute.} Embedding all extracted opinions with OpenAI \texttt{text-embedding-3-small} costs roughly \$5--\$20. The released metric JSONs allow readers to inspect and reproduce the reported numerical summaries without rerunning generation, extraction, or embedding. All metric computations and statistical tests run on a single CPU in under one hour for the full evaluation set.

\section{Released Artifacts}
\label{app:artifacts}

This appendix documents the code artifact released alongside this paper, where to find it, and what it contains. The release supports the contribution claim of a reusable evaluation protocol; it is not a stand-alone dataset submission.

\paragraph{Repository structure.} The codebase is hosted at \url{https://github.com/MeYQY/more-is-not-more-neurips2026}. The repository top-level layout is:
\begin{quote}\ttfamily\small
src/modbench/      \# pipeline modules (runner, extractor, embedding, metrics, stats)\\
scripts/           \# entry points (run\_experiment, extract, embed, compute\_metrics, plot\_figures)\\
configs/           \# base.yaml, conditions/*.yaml, personas.yaml, extraction.yaml\\
data/dataset/      \# tasks.csv (100 question IDs + metadata)\\
results/           \# aggregated per-condition metric JSONs and summary outputs\\
validation/        \# extraction-audit artifacts (see below)\\
rebuttal/          \# human evaluation, robustness, Multi-Agent prompt sensitivity (App.~\ref{app:human-eval}--\ref{app:ma-prompt})\\
README.md          \# reproduction quickstart\\
pyproject.toml     \# locked dependencies
\end{quote}

The \texttt{validation/} directory contains every artifact cited from Appendix~\ref{app:extraction}: \texttt{human\_annotations\_30.json} (30-record human precision pass), \texttt{sonnet\_annotations.json} and \texttt{sonnet\_chunk\_*.json} (150-record LLM precision pass), \texttt{cross\_extractor/report.json} (Spearman / Pearson / per-architecture Jaccard against Kimi K2.5), \texttt{social\_language\_analysis.json} (per-architecture social-utterance frequency cited in the density check), and \texttt{records\_data.json} and \texttt{sample\_manifest.json} (record-level metadata for the audited samples).

\paragraph{Submitted artifact.}
The submitted codebase contains the 100-question ID list with metadata, the 19 primary condition configs, the two ablation configs, the full persona and prompt definitions, the validation annotations, and aggregated per-condition metric JSONs. This is sufficient to inspect the experimental design, audit the evaluation implementation, and reproduce the reported numerical summaries without re-running the full LLM generation stage. We do not submit a separate dataset artifact containing the full raw LLM responses or full extracted-opinion corpus, because the paper's contribution is the evaluation protocol and empirical audit rather than a reusable generated-response dataset. The repository ships the generation, extraction, embedding, and metric scripts so these intermediate artifacts can be regenerated from the included task and condition manifests, subject to the API costs in Appendix~\ref{app:compute}.

\paragraph{Licensing chain.} The 100-question set derives from WildChat-1M~\citep{zhao2024wildchat}, which is released under ODC-BY 1.0 and re-distributable with attribution. We redistribute the 100-question metadata subset with the same attribution chain. The codebase is released under CC BY 4.0. LLM-generated responses are subject to the originating provider's terms-of-service and are not submitted as a standalone dataset.

\paragraph{Reproduction quickstart.} A minimal single-condition replication proceeds in four commands after \texttt{pip install -e .}:
\begin{quote}\ttfamily\small
python scripts/run\_experiment.py\\
\quad configs/conditions/independent/persona\_pro.yaml\\
python scripts/extract.py independent-pro\\
python scripts/embed.py --embedder openai-3-small --conditions independent-pro\\
python scripts/compute\_metrics.py --embedder openai-3-small\\
\quad --conditions independent-pro
\end{quote}
The same pipeline applies to the 19 primary conditions and the seven model configs (only the base config is swapped; GPT-5.4-mini has the partial coverage noted in Appendix~\ref{app:models}). Pre-computed metric JSONs (\texttt{results/metrics/}) are included in the repository so that figure-regeneration scripts (\texttt{scripts/plot\_figures.py}) run in seconds without any API calls.

\section{Ethics, Broader Impacts, and LLM Usage}
\label{app:ethics}

This appendix addresses NeurIPS Checklist Q9, Q10, Q11, and Q16.

\paragraph{Data sourcing and de-identification.} The 100 evaluation questions are sampled from the publicly released WildChat-1M corpus~\citep{zhao2024wildchat}, which already removes user account identifiers and PII at the source. We additionally manually reviewed all 100 selected queries and confirmed that none contain personal names, contact information, addresses, or other identifying content; any such item would have been excluded during selection. We retain only the query text and the original WildChat conversation ID (an opaque hash). The only data collected from human participants is the blinded Prolific evaluation in Appendix~\ref{app:human-eval}, a minimal-risk comparison of AI-generated texts conducted with informed consent; no IRB review was obtained, and only de-identified judgments are released.

\paragraph{Broader impacts.} The work has dual-use character. On the constructive side, a controlled understanding of LLM opinion diversity supports legitimate uses of LLM simulation: synthetic survey research, focus-group prototyping, public-opinion modeling, and the design of more representative conversational agents---all of which benefit from knowing which interventions actually deliver coverage and which do not. On the misuse side, the same techniques can be applied to inflate the apparent diversity of synthetic opinion generators used for astroturfing or coordinated influence operations.

We argue that publishing this audit asymmetrically benefits the defensive side. The misuse direction was already feasible---practitioners building synthetic-opinion pipelines today combine personas, multi-agent frameworks, and temperature changes without disciplined evaluation, and our results show several of these recipes (high temperature, generic ``consider diverse perspectives'' instructions, demographic cueing alone) deliver near-zero gain. The auditing direction, by contrast, requires the systematic factorial decomposition we contribute: without it, evaluators of synthetic-opinion systems lack the ground truth to challenge over-claims about diversity. We therefore expect the net effect to favor evaluation and accountability over uplift.

We do not release any new model, generation system, production-ready data-generation pipeline, or standalone generated-response dataset. The released artifact is an evaluation protocol with task metadata, configs, validation files, and metric outputs---components useful to scrutinize systems that already exist, not to build new ones.

\paragraph{LLM usage in the methodology.} LLMs are both the object of study (the seven generation models in Table~\ref{tab:models}) and a tool in the pipeline (DeepSeek v3.2 as the opinion extractor, with Kimi K2.5 as the cross-extractor probe). All such usage is methodologically central and is documented at the prompt and parameter level in Appendix~\ref{app:prompts}. LLM-based assistants were also used for code development and for polishing the writing (grammar, phrasing, table formatting). The author reviewed all code, analyses, and text and takes full responsibility for the paper's content.

%% file: tables/T2_questions.tex
\begin{table}[!htbp]
\centering
\small
\caption{Subcategory distribution of the 100 curated questions. ``Mean query length'' is the number of characters in the user's query text, not including any model response.}
\label{tab:questions}
\begin{tabular}{lrr}
\toprule
Subcategory & Count & Mean query length (chars) \\
\midrule
Analytical and Interpretive Questions & 20 & 119 \\
Speculative and Hypothetical Scenarios & 18 & 82 \\
Controversial Questions & 12 & 100 \\
Value-Laden Questions & 12 & 86 \\
Opinion-Based Questions & 10 & 58 \\
Personal Advice & 8 & 75 \\
Philosophical Questions & 8 & 56 \\
Abstract Conceptual Questions & 5 & 66 \\
Ideation and Brainstorming & 4 & 47 \\
Recommendations & 3 & 50 \\
\midrule
\textbf{Total} & \textbf{100} & \textbf{84} \\
\bottomrule
\end{tabular}
\end{table}

%% file: tables/T3_personas_demo.tex
\begin{table}[!htbp]
\centering
\small
\caption{Demographic distribution of the 20 personas. Each persona is described at four depth levels (Role, Basic, Mid, Pro); attributes shown here are extracted from the Mid level. Geographic bucket follows the U.S.\ Census four-region scheme.}
\label{tab:personas-demo}
\begin{tabular}{lr}
\toprule
Attribute & Count (\%) \\
\midrule
\multicolumn{2}{l}{\textit{Age}} \\
\quad 18--29 & 5 (25\%) \\
\quad 30--44 & 8 (40\%) \\
\quad 45--59 & 4 (20\%) \\
\quad 60+ & 3 (15\%) \\
\midrule
\multicolumn{2}{l}{\textit{Gender}} \\
\quad Female & 11 (55\%) \\
\quad Male & 9 (45\%) \\
\midrule
\multicolumn{2}{l}{\textit{Race}} \\
\quad White & 6 (30\%) \\
\quad Black & 5 (25\%) \\
\quad Hispanic & 4 (20\%) \\
\quad Asian American & 4 (20\%) \\
\quad Somali American & 1 (5\%) \\
\midrule
\multicolumn{2}{l}{\textit{Region}} \\
\quad Northeast & 4 (20\%) \\
\quad Midwest & 4 (20\%) \\
\quad South & 8 (40\%) \\
\quad West & 4 (20\%) \\
\midrule
\multicolumn{2}{l}{\textit{Political leaning}} \\
\quad Democrat & 6 (30\%) \\
\quad Republican & 6 (30\%) \\
\quad Liberal/Progressive & 5 (25\%) \\
\quad Libertarian & 1 (5\%) \\
\quad Moderate & 1 (5\%) \\
\quad Not politically engaged & 1 (5\%) \\
\bottomrule
\end{tabular}
\end{table}

%% file: tables/T1_models.tex
\begin{table}[!htbp]
\centering
\small
\caption{The seven chat models evaluated. Compact result tables abbreviate GPT-5.4-mini as GPT-mini. Models without explicit notes use the unified generation parameters of $T=0.7$, top-$p = 1.0$, $\max_{\text{tokens}} = 4096$, no frequency or presence penalty. Kimi K2.5 is run in non-thinking mode (the only mode it allows top-$p < 1$); Grok and Qwen are run with their reasoning modes explicitly disabled to match the other chat models.}
\label{tab:models}
\resizebox{\textwidth}{!}{\begin{tabular}{lllll}
\toprule
Model (paper) & Provider & Version identifier & Endpoint host & Notes \\
\midrule
DeepSeek & DeepSeek & \texttt{deepseek-chat} & \texttt{api.deepseek.com} & --- \\
GPT-5.4-mini & OpenAI & \texttt{openai/gpt-5.4-mini} & \texttt{openrouter.ai} & --- \\
Gemini & Google & \texttt{google/gemini-3-flash-preview} & \texttt{openrouter.ai} & --- \\
Kimi & Moonshot & \texttt{kimi-k2.5} & \texttt{api.moonshot.ai} & thinking off, $T{=}0.6$, $\text{top-}p{=}0.95$ \\
Llama & Meta & \texttt{meta-llama/llama-4-maverick} & \texttt{openrouter.ai} & --- \\
Grok & xAI & \texttt{x-ai/grok-4.1-fast} & \texttt{openrouter.ai} & reasoning off \\
Qwen & Alibaba & \texttt{qwen/qwen3.6-plus} & \texttt{openrouter.ai} & reasoning off \\
\bottomrule
\end{tabular}}
\end{table}

%% file: tables/T4_conditions.tex
\begin{table}[!htbp]
\centering
\footnotesize
\caption{The 19 primary experimental conditions plus 2 persona-pool-size ablation variants (\texttt{persona\_5}, \texttt{mt\_pro\_5}). ``Personas'' is the number of distinct identity profiles drawn from the 20-persona pool. ``Rounds'' is the number of conversation turns (Single-Call: one independent call per persona; Multi-Turn: three-turn self-prompted conversation per persona; Multi-Agent: three-round group discussion among matched personas). ``Resp/Q'' is the resulting number of responses per question.}
\label{tab:conditions}
\begin{tabular}{lllrrr}
\toprule
Condition (config name) & Architecture & Persona depth & Personas & Rounds & Resp/Q \\
\midrule
\texttt{dimension} & Single-Call & None & 0 & 1 & 20 \\
\texttt{enhanced} & Single-Call & None & 0 & 1 & 20 \\
\texttt{persona\_5} & Single-Call & Pro & 5 & 1 & 5 \\
\texttt{persona\_basic} & Single-Call & Basic & 20 & 1 & 20 \\
\texttt{persona\_mid} & Single-Call & Mid & 20 & 1 & 20 \\
\texttt{persona\_pro} & Single-Call & Pro & 20 & 1 & 20 \\
\texttt{random\_sys} & Single-Call & None & 5 traits & 1 & 20 \\
\texttt{role} & Single-Call & Role & 20 & 1 & 20 \\
\texttt{single} & Single-Call & None & 0 & 1 & 20 \\
\texttt{temp\_high} & Single-Call & None & 0 & 1 & 20 \\
\midrule
\texttt{mt\_basic} & Multi-Turn & Basic & 20 & 3 & 60 \\
\texttt{mt\_mid} & Multi-Turn & Mid & 20 & 3 & 60 \\
\texttt{mt\_none} & Multi-Turn & None & 5 anon. & 3 & 15 \\
\texttt{mt\_pro} & Multi-Turn & Pro & 20 & 3 & 60 \\
\texttt{mt\_pro\_5} & Multi-Turn & Pro & 5 & 3 & 15 \\
\texttt{mt\_role} & Multi-Turn & Role & 20 & 3 & 60 \\
\midrule
\texttt{minimal\_bare} & Multi-Agent & None & 5 anon. & 3 & 15 \\
\texttt{minimal\_basic} & Multi-Agent & Basic & 5 & 3 & 15 \\
\texttt{minimal\_mid} & Multi-Agent & Mid & 5 & 3 & 15 \\
\texttt{minimal\_pro} & Multi-Agent & Pro & 5 & 3 & 15 \\
\texttt{minimal\_role} & Multi-Agent & Role & 5 & 3 & 15 \\
\bottomrule
\end{tabular}
\end{table}

%% file: tables/T14_extraction_precision.tex
\begin{table}[!htbp]
\centering
\small
\caption{Extraction-fidelity audit. Each row is a precision rate computed on a stratified sample of (record, opinion) pairs drawn equally from the three architectures. ``Strict'' counts only judgments marked Yes; ``Lenient'' additionally counts Partial. The two passes use different annotators (the author and an independent LLM judge) and different sample sizes.}
\label{tab:extraction-precision}
\begin{tabular}{lrrrrr}
\toprule
Architecture & Records & Opinions & Strict precision & Lenient precision & Partial+No (n) \\
\midrule
\multicolumn{6}{l}{\textit{Human pass (author, 398 stratified opinions)}} \\
\quad Independent & 10 & 142 & 99.3\% & 99.3\% & 1 \\
\quad Sequential & 10 & 116 & 97.4\% & 98.3\% & 3 \\
\quad Interactive & 10 & 140 & 97.9\% & 100.0\% & 3 \\
\quad \textbf{All} & 30 & 398 & \textbf{98.2\%} & 99.2\% & 7 \\
\midrule
\multicolumn{6}{l}{\textit{LLM pass (independent judge, 150 records)}} \\
\quad Independent & 50 & 619 & 98.2\% & 100.0\% & 11 \\
\quad Sequential & 50 & 611 & 98.9\% & 99.2\% & 7 \\
\quad Interactive & 50 & 621 & 98.1\% & 99.7\% & 12 \\
\quad \textbf{All} & 150 & 1851 & \textbf{98.4\%} & 99.6\% & 30 \\
\bottomrule
\end{tabular}
\end{table}

%% file: tables/T15_cross_extractor.tex
\begin{table}[!htbp]
\centering
\small
\caption{Extractor independence. We re-extract a fixed 20-task subset on DeepSeek under the 19 primary conditions plus one ablation variant (the supplementary \texttt{mt\_pro\_5} is omitted because it shares the same 5 personas as \texttt{persona\_5} and thus would not produce a distinct ranking signal) using Kimi K2.5 as an alternative extractor (vs.\ DeepSeek v3.2 in the main pipeline). Condition-level MPD ranks are highly preserved (Spearman $\rho = 0.986$, $p < 10^{-15}$) despite substantial stylistic differences in individual record extractions (mean record-level Jaccard $= 0.57$). The per-architecture record Jaccard breakdown rules out a Kimi-vs-DeepSeek bias toward any single architecture.}
\label{tab:cross-extractor}
\begin{tabular}{lr}
\toprule
Quantity & Value \\
\midrule
\multicolumn{2}{l}{\textit{Condition-level agreement (across 20 conditions)}} \\
\quad Spearman $\rho$ & 0.9865 ($p=1.4e-15$) \\
\quad Pearson $r$ & 0.9945 \\
\midrule
\multicolumn{2}{l}{\textit{Record-level Jaccard ($n=142$)}} \\
\quad Mean (std) & 0.572 (0.245) \\
\quad Median & 0.587 \\
\quad Range & [0.00, 1.00] \\
\midrule
\multicolumn{2}{l}{\textit{Per-architecture mean Jaccard}} \\
\quad Independent & 0.608 ($n=48$) \\
\quad Sequential & 0.604 ($n=48$) \\
\quad Interactive & 0.499 ($n=46$) \\
\midrule
\multicolumn{2}{l}{\textit{Granularity (mean opinions per record)}} \\
\quad DeepSeek extractor & 12.34 \\
\quad Kimi K2.5 extractor & 13.33 \\
\bottomrule
\end{tabular}
\end{table}

%% file: tables/T10_embedder.tex
\begin{table}[!htbp]
\centering
\small
\caption{Embedding-model sensitivity. Spearman rank correlation between MPD rankings under text-embedding-3-small and two alternative embedders, computed at the task level (paired across 100 questions) and condition level (across the 19 primary conditions plus 2 ablation variants per model).}
\label{tab:embedder}
\begin{tabular}{lrr}
\toprule
Embedder $\times$ Model & Task-level $\rho$ & Condition-level $\rho$ \\
\midrule
BGE-M3 $\times$ DeepSeek & 0.7488 & 0.9401 \\
Qwen3-Embedding-8B $\times$ DeepSeek & 0.7843 & 0.9422 \\
BGE-M3 $\times$ Gemini & 0.6716 & 0.8824 \\
Qwen3-Embedding-8B $\times$ Gemini & 0.7307 & 0.9649 \\
\bottomrule
\end{tabular}
\end{table}

%% file: tables/T11_threshold.tex
\begin{table}[!htbp]
\centering
\small
\caption{Clustering-threshold sensitivity. Mean Cluster Count (CC) per task on DeepSeek across seven representative conditions and five thresholds $\tau \in \{0.55, 0.60, 0.65, 0.70, 0.75\}$. Mean across 100 tasks (no rarefaction; threshold sensitivity is independent of rarefaction since both rely on the same underlying linkage). Headline orderings are preserved across all five thresholds: Multi-Turn $\times$ Pro is the most diverse condition at every $\tau$; because these counts are not rarefied, they also reflect opinion volume (rarefied comparisons are in Table~\ref{tab:alpha-cc-rare}). Adjacent thresholds yield Spearman $\rho \geq 0.93$; rank-correlation between the most distant thresholds ($\tau{=}0.55$ and $\tau{=}0.75$) drops to $\rho=0.54$, reflecting that a high $\tau$ counts each opinion as nearly its own cluster and thus makes CC track raw opinion volume rather than category richness.}
\label{tab:threshold-sensitivity}
\begin{tabular}{lrrrrr}
\toprule
Condition & $\tau{=}0.55$ & $\tau{=}0.60$ & $\tau{=}0.65$ & $\tau{=}0.70$ & $\tau{=}0.75$ \\
\midrule
Single-Call $\times$ None & 61.3 & 85.2 & 114.1 & 146.9 & 182.8 \\
Single-Call $\times$ Role & 71.6 & 89.2 & 106.6 & 121.4 & 133.6 \\
Single-Call $\times$ Pro & 102.9 & 121.7 & 139.2 & 154.1 & 166.3 \\
Multi-Turn $\times$ None & 137.5 & 178.3 & 217.8 & 251.1 & 276.7 \\
Multi-Turn $\times$ Pro & 392.9 & 467.4 & 529.5 & 577.1 & 609.4 \\
Multi-Agent $\times$ None & 76.5 & 104.9 & 134.2 & 159.9 & 182.5 \\
Multi-Agent $\times$ Pro & 101.8 & 118.2 & 131.8 & 142.0 & 148.1 \\
\bottomrule
\end{tabular}
\end{table}

%% file: tables/T19_baseline.tex
\begin{table}[!htbp]
\centering
\small
\caption{$\beta$-diversity baseline calibration on DeepSeek. Split-half divides the None condition's opinions in half; None vs.\ High-Temperature compares two conditions with near-identical $\alpha$-diversity (MPD difference $<0.01$).}
\label{tab:baseline}
\begin{tabular}{lrr}
\toprule
Comparison & $\beta$-VS & UCR \\
\midrule
Split-half of None (random partition) & 4.01 & 29.3\% \\
None vs.\ High Temperature ($\Delta$MPD $<$ 0.01) & 3.28 & 24.7\% \\
\bottomrule
\end{tabular}
\end{table}

%% file: tables/T5a_mpd.tex
\begin{table}[!htbp]
\centering
\small
\caption{Mean Pairwise Distance (MPD) per condition. Five persona depths within each architecture; 7 models. MPD is sample-size invariant. Em-dash (---) indicates unrun condition for GPT-mini.}
\label{tab:alpha-mpd}
\begin{tabular}{llrrrrrrr}
\toprule
Architecture & Persona & DeepSeek & GPT-mini & Gemini & Kimi & Llama & Grok & Qwen \\
\midrule
Single-Call & None & 0.5979 & 0.5774 & 0.6290 & 0.6230 & 0.5353 & 0.6257 & 0.6065 \\
 & Role & 0.6661 & 0.6102 & 0.7134 & 0.7079 & 0.5763 & 0.6536 & 0.6829 \\
 & Basic & 0.6847 & 0.5799 & 0.6943 & 0.6826 & 0.5945 & 0.6511 & 0.6761 \\
 & Mid & 0.6757 & 0.5887 & 0.7013 & 0.6862 & 0.5974 & 0.6577 & 0.6811 \\
 & Pro & 0.7022 & 0.5967 & 0.7224 & 0.7052 & 0.6156 & 0.6649 & 0.6952 \\
\midrule
Multi-Turn & None & 0.6584 & --- & 0.6727 & 0.6852 & 0.5830 & 0.6808 & 0.6525 \\
 & Role & 0.7193 & --- & 0.7417 & 0.7689 & 0.6559 & 0.7221 & 0.7467 \\
 & Basic & 0.7251 & --- & 0.7270 & 0.7364 & 0.6394 & 0.7129 & 0.7144 \\
 & Mid & 0.7183 & --- & 0.7326 & 0.7405 & 0.6455 & 0.7197 & 0.7225 \\
 & Pro & 0.7429 & 0.6670 & 0.7464 & 0.7519 & 0.6682 & 0.7248 & 0.7474 \\
\midrule
Multi-Agent & None & 0.6144 & --- & 0.6497 & 0.6865 & 0.5586 & 0.6673 & 0.6727 \\
 & Role & 0.6763 & --- & 0.7328 & 0.7580 & 0.6161 & 0.7171 & 0.7540 \\
 & Basic & 0.6859 & --- & 0.7228 & 0.7397 & 0.5926 & 0.6954 & 0.7453 \\
 & Mid & 0.6681 & --- & 0.7222 & 0.7445 & 0.6030 & 0.6993 & 0.7402 \\
 & Pro & 0.7166 & 0.6363 & 0.7414 & 0.7551 & 0.6277 & 0.7069 & 0.7552 \\
\bottomrule
\end{tabular}
\end{table}

%% file: tables/T5b_cc.tex
\begin{table}[!htbp]
\centering
\small
\caption{Cluster Count (CC, raw) per condition; agglomerative clustering with cosine threshold $\tau=0.65$. CC is sensitive to opinion count $N$; rarefied counterpart in Table~\ref{tab:alpha-cc-rare}.}
\label{tab:alpha-cc-raw}
\begin{tabular}{llrrrrrrr}
\toprule
Architecture & Persona & DeepSeek & GPT-mini & Gemini & Kimi & Llama & Grok & Qwen \\
\midrule
Single-Call & None & 114.1 & 90.6 & 139.1 & 99.4 & 75.4 & 141.0 & 117.7 \\
 & Role & 106.6 & 111.9 & 203.0 & 123.8 & 67.4 & 150.6 & 159.9 \\
 & Basic & 112.1 & 87.8 & 155.2 & 107.5 & 72.3 & 151.0 & 166.0 \\
 & Mid & 105.7 & 97.5 & 160.7 & 114.3 & 74.4 & 153.9 & 173.4 \\
 & Pro & 139.2 & 103.3 & 170.3 & 118.1 & 79.0 & 158.1 & 174.5 \\
\midrule
Multi-Turn & None & 217.8 & --- & 217.8 & 152.1 & 80.8 & 183.6 & 186.6 \\
 & Role & 734.3 & --- & 724.8 & 565.6 & 282.4 & 680.8 & 809.1 \\
 & Basic & 424.4 & --- & 777.9 & 403.3 & 223.8 & 591.0 & 627.9 \\
 & Mid & 413.7 & --- & 587.7 & 417.1 & 222.8 & 599.0 & 669.9 \\
 & Pro & 529.5 & 412.2 & 604.5 & 401.2 & 224.0 & 603.5 & 973.1 \\
\midrule
Multi-Agent & None & 134.2 & --- & 158.9 & 250.1 & 46.1 & 130.5 & 226.3 \\
 & Role & 105.1 & --- & 166.7 & 178.5 & 62.2 & 153.0 & 223.6 \\
 & Basic & 102.7 & --- & 138.1 & 135.4 & 46.5 & 121.3 & 190.3 \\
 & Mid & 91.9 & --- & 143.1 & 146.5 & 49.6 & 123.7 & 216.3 \\
 & Pro & 131.8 & 117.3 & 154.2 & 139.2 & 55.3 & 120.0 & 220.2 \\
\bottomrule
\end{tabular}
\end{table}

%% file: tables/T5c_vs.tex
\begin{table}[!htbp]
\centering
\small
\caption{Vendi Score (VS, raw) per condition. VS is sensitive to opinion count $N$; rarefied counterpart in Table~\ref{tab:alpha-vs-rare}.}
\label{tab:alpha-vs-raw}
\begin{tabular}{llrrrrrrr}
\toprule
Architecture & Persona & DeepSeek & GPT-mini & Gemini & Kimi & Llama & Grok & Qwen \\
\midrule
Single-Call & None & 26.13 & 21.64 & 31.97 & 26.86 & 18.62 & 31.21 & 27.01 \\
 & Role & 30.47 & 26.19 & 49.84 & 38.73 & 20.02 & 35.65 & 40.27 \\
 & Basic & 33.48 & 21.72 & 41.24 & 33.96 & 21.38 & 35.80 & 39.67 \\
 & Mid & 31.71 & 23.31 & 42.67 & 35.49 & 21.89 & 36.74 & 41.25 \\
 & Pro & 39.99 & 24.61 & 47.58 & 38.07 & 23.77 & 38.12 & 43.39 \\
\midrule
Multi-Turn & None & 41.80 & --- & 44.55 & 39.80 & 22.34 & 43.30 & 38.38 \\
 & Role & 74.06 & --- & 86.87 & 94.76 & 43.06 & 79.98 & 89.02 \\
 & Basic & 66.80 & --- & 79.98 & 72.14 & 37.53 & 73.77 & 72.07 \\
 & Mid & 64.39 & --- & 78.48 & 74.90 & 38.06 & 76.47 & 76.40 \\
 & Pro & 79.08 & 49.60 & 85.35 & 77.65 & 41.80 & 78.94 & 91.53 \\
\midrule
Multi-Agent & None & 29.27 & --- & 36.06 & 47.78 & 16.33 & 35.29 & 43.73 \\
 & Role & 31.34 & --- & 48.37 & 56.40 & 20.92 & 44.49 & 59.40 \\
 & Basic & 33.02 & --- & 43.43 & 45.97 & 17.37 & 37.26 & 54.38 \\
 & Mid & 29.53 & --- & 44.02 & 48.94 & 18.25 & 38.31 & 56.99 \\
 & Pro & 41.62 & 29.22 & 49.03 & 49.52 & 20.27 & 38.63 & 60.87 \\
\bottomrule
\end{tabular}
\end{table}

%% file: tables/T6a_cc_rare.tex
\begin{table}[!htbp]
\centering
\small
\caption{Rarefied Cluster Count (CC). Within each comparison group, all conditions are randomly downsampled to the minimum opinion count, repeated 30 times, and averaged. Removes the volume artifact in Table~\ref{tab:alpha-cc-raw}.}
\label{tab:alpha-cc-rare}
\begin{tabular}{llrrrrrrr}
\toprule
Architecture & Persona & DeepSeek & GPT-mini & Gemini & Kimi & Llama & Grok & Qwen \\
\midrule
Single-Call & None & 59.7 & 78.0 & 93.3 & 67.9 & 40.3 & 111.5 & 79.0 \\
 & Role & 79.9 & 96.4 & 159.0 & 97.7 & 54.2 & 124.8 & 125.0 \\
 & Basic & 84.7 & 73.9 & 132.1 & 88.2 & 61.1 & 128.6 & 127.2 \\
 & Mid & 80.1 & 84.7 & 135.1 & 94.3 & 61.5 & 126.1 & 132.9 \\
 & Pro & 95.8 & 88.0 & 136.6 & 91.1 & 62.6 & 132.6 & 127.4 \\
\midrule
Multi-Turn & None & 153.3 & --- & 148.3 & 102.8 & 60.3 & 123.5 & 129.1 \\
 & Role & 258.2 & --- & 273.2 & 179.9 & 119.4 & 206.1 & 266.2 \\
 & Basic & 245.8 & --- & 254.4 & 164.7 & 110.4 & 198.0 & 242.6 \\
 & Mid & 249.0 & --- & 263.0 & 167.6 & 111.8 & 201.1 & 252.0 \\
 & Pro & 267.9 & --- & 271.5 & 170.2 & 119.7 & 204.6 & 264.8 \\
\midrule
Multi-Agent & None & 75.0 & --- & 100.5 & 110.9 & 35.5 & 87.6 & 133.4 \\
 & Role & 80.5 & --- & 128.3 & 120.2 & 47.8 & 105.1 & 155.7 \\
 & Basic & 79.7 & --- & 111.1 & 104.3 & 38.5 & 93.7 & 137.1 \\
 & Mid & 72.3 & --- & 117.4 & 111.6 & 40.7 & 96.9 & 152.6 \\
 & Pro & 93.2 & --- & 124.2 & 104.4 & 44.8 & 92.0 & 155.5 \\
\bottomrule
\end{tabular}
\end{table}

%% file: tables/T6b_vs_rare.tex
\begin{table}[!htbp]
\centering
\small
\caption{Rarefied Vendi Score (VS). Same rarefaction procedure as Table~\ref{tab:alpha-cc-rare}.}
\label{tab:alpha-vs-rare}
\begin{tabular}{llrrrrrrr}
\toprule
Architecture & Persona & DeepSeek & GPT-mini & Gemini & Kimi & Llama & Grok & Qwen \\
\midrule
Single-Call & None & 20.06 & 20.44 & 27.49 & 22.80 & 14.84 & 28.26 & 23.45 \\
 & Role & 25.69 & 24.14 & 44.59 & 33.58 & 17.75 & 32.19 & 35.37 \\
 & Basic & 28.39 & 20.03 & 37.80 & 30.02 & 19.46 & 32.98 & 34.78 \\
 & Mid & 27.22 & 22.02 & 38.71 & 31.58 & 19.66 & 33.34 & 36.22 \\
 & Pro & 32.86 & 22.96 & 41.62 & 32.29 & 20.79 & 34.66 & 36.61 \\
\midrule
Multi-Turn & None & 34.91 & --- & 36.16 & 31.14 & 18.80 & 34.16 & 31.53 \\
 & Role & 55.42 & --- & 64.08 & 58.94 & 31.60 & 53.12 & 62.69 \\
 & Basic & 56.11 & --- & 58.60 & 49.71 & 29.21 & 50.84 & 53.41 \\
 & Mid & 54.81 & --- & 60.63 & 51.05 & 29.65 & 52.50 & 55.88 \\
 & Pro & 63.46 & --- & 65.54 & 53.63 & 32.88 & 54.07 & 63.24 \\
\midrule
Multi-Agent & None & 22.92 & --- & 29.77 & 34.38 & 14.37 & 28.84 & 35.85 \\
 & Role & 27.23 & --- & 42.29 & 45.74 & 18.20 & 36.18 & 49.43 \\
 & Basic & 28.43 & --- & 37.58 & 39.11 & 15.53 & 32.28 & 45.05 \\
 & Mid & 25.37 & --- & 39.17 & 41.85 & 16.35 & 33.19 & 47.66 \\
 & Pro & 34.59 & --- & 43.26 & 41.32 & 17.97 & 32.74 & 50.86 \\
\bottomrule
\end{tabular}
\end{table}

%% file: tables/T7_pairwise_mpd.tex
\begin{table}[!htbp]
\centering
\small
\caption{Pairwise comparisons on MPD: matched-pairs Cliff's $\delta$ (paired rank-biserial form). Categorical thresholds follow~\citet{romano2006}: neg ($|\delta|<0.147$), sma ($<0.33$), med ($<0.474$), lar ($\geq 0.474$). Asterisk ($*$) denotes $p_{\textsc{fdr}} < 0.05$ after Benjamini--Hochberg correction. Em-dash indicates GPT-mini did not run that comparison. Positive values indicate that the first-listed condition is higher.}
\label{tab:pair-mpd}
\begin{tabular}{llrrrrrrr}
\toprule
ID & Comparison & DeepSeek & GPT-mini & Gemini & Kimi & Llama & Grok & Qwen \\
\midrule
A1 & None$\to$Role & $-0.82{}^{*}$ & $-0.72{}^{*}$ & $-1.00{}^{*}$ & $-1.00{}^{*}$ & $-0.76{}^{*}$ & $-0.62{}^{*}$ & $-0.90{}^{*}$ \\
A2 & Role$\to$Basic & $-0.42{}^{*}$ & $+0.72{}^{*}$ & $+0.62{}^{*}$ & $+0.60{}^{*}$ & $-0.46{}^{*}$ & $-0.08$ & $+0.14$ \\
A3 & Basic$\to$Mid & $+0.40{}^{*}$ & $-0.34{}^{*}$ & $-0.38{}^{*}$ & $-0.18$ & $-0.06$ & $-0.28{}^{*}$ & $-0.06$ \\
A4 & Mid$\to$Pro & $-0.84{}^{*}$ & $-0.28{}^{*}$ & $-0.92{}^{*}$ & $-0.54{}^{*}$ & $-0.50{}^{*}$ & $-0.36{}^{*}$ & $-0.54{}^{*}$ \\
A5 & None$\to$Pro & $-0.98{}^{*}$ & $-0.50{}^{*}$ & $-0.94{}^{*}$ & $-0.94{}^{*}$ & $-0.84{}^{*}$ & $-0.86{}^{*}$ & $-0.94{}^{*}$ \\
B1 & vs Enhanced & $-0.18{}^{*}$ & --- & $-0.08$ & $-0.39{}^{*}$ & $+0.09$ & $-0.28{}^{*}$ & $-0.41{}^{*}$ \\
B2 & vs TempHigh & $-0.06$ & --- & $-0.12$ & --- & $-0.22{}^{*}$ & $-0.18$ & $-0.19{}^{*}$ \\
B3 & vs Dimension & $+0.18$ & --- & $-0.10$ & $-0.15$ & $+0.04$ & $+0.00$ & $-0.39{}^{*}$ \\
B4 & vs RandomSys & $-0.44{}^{*}$ & --- & $-0.88{}^{*}$ & $-0.84{}^{*}$ & $-0.30{}^{*}$ & $-0.32{}^{*}$ & $-0.76{}^{*}$ \\
C1 & Indep$\to$Seq @None & $-0.96{}^{*}$ & --- & $-0.86{}^{*}$ & $-0.92{}^{*}$ & $-0.70{}^{*}$ & $-0.96{}^{*}$ & $-0.82{}^{*}$ \\
C2 & Indep$\to$Int @None & $-0.30{}^{*}$ & --- & $-0.46{}^{*}$ & $-0.86{}^{*}$ & $-0.24{}^{*}$ & $-0.70{}^{*}$ & $-0.84{}^{*}$ \\
C3 & Seq vs Int @None & $+0.86{}^{*}$ & --- & $+0.66{}^{*}$ & $-0.05$ & $+0.49{}^{*}$ & $+0.24{}^{*}$ & $-0.40{}^{*}$ \\
C4 & Indep$\to$Seq @Pro & $-0.92{}^{*}$ & $-0.94{}^{*}$ & $-0.82{}^{*}$ & $-0.98{}^{*}$ & $-0.84{}^{*}$ & $-1.00{}^{*}$ & $-0.96{}^{*}$ \\
C5 & Indep$\to$Int @Pro & $-0.22{}^{*}$ & $-0.62{}^{*}$ & $-0.58{}^{*}$ & $-0.84{}^{*}$ & $-0.04$ & $-0.84{}^{*}$ & $-0.88{}^{*}$ \\
C6 & Seq vs Int @Pro & $+0.64{}^{*}$ & $+0.68{}^{*}$ & $+0.18{}^{*}$ & $-0.05$ & $+0.80{}^{*}$ & $+0.54{}^{*}$ & $-0.18{}^{*}$ \\
C7 & Indep$\to$Seq @Role & $-0.96{}^{*}$ & --- & $-0.80{}^{*}$ & $-0.98{}^{*}$ & $-0.90{}^{*}$ & $-0.98{}^{*}$ & $-0.98{}^{*}$ \\
C8 & Indep$\to$Int @Role & $-0.28{}^{*}$ & --- & $-0.54{}^{*}$ & $-0.82{}^{*}$ & $-0.52{}^{*}$ & $-0.92{}^{*}$ & $-0.90{}^{*}$ \\
C9 & Seq vs Int @Role & $+0.94{}^{*}$ & --- & $+0.26{}^{*}$ & $+0.46{}^{*}$ & $+0.76{}^{*}$ & $+0.10$ & $-0.16{}^{*}$ \\
D0 & Seq: None$\to$Role & $-0.98{}^{*}$ & --- & $-1.00{}^{*}$ & $-1.00{}^{*}$ & $-0.94{}^{*}$ & $-0.98{}^{*}$ & $-1.00{}^{*}$ \\
D1 & Seq: None$\to$Basic & $-0.96{}^{*}$ & --- & $-0.98{}^{*}$ & $-0.98{}^{*}$ & $-0.88{}^{*}$ & $-0.92{}^{*}$ & $-1.00{}^{*}$ \\
D2 & Seq: Basic$\to$Mid & $+0.44{}^{*}$ & --- & $-0.48{}^{*}$ & $-0.26{}^{*}$ & $-0.20{}^{*}$ & $-0.54{}^{*}$ & $-0.44{}^{*}$ \\
D3 & Seq: Mid$\to$Pro & $-0.96{}^{*}$ & --- & $-0.94{}^{*}$ & $-0.58{}^{*}$ & $-0.78{}^{*}$ & $-0.37{}^{*}$ & $-0.78{}^{*}$ \\
D4 & Seq: None$\to$Pro & $-1.00{}^{*}$ & --- & $-1.00{}^{*}$ & $-1.00{}^{*}$ & $-1.00{}^{*}$ & $-0.96{}^{*}$ & $-1.00{}^{*}$ \\
E0 & Int: None$\to$Role & $-0.90{}^{*}$ & --- & $-0.98{}^{*}$ & $-1.00{}^{*}$ & $-0.72{}^{*}$ & $-0.94{}^{*}$ & $-1.00{}^{*}$ \\
E1 & Int: None$\to$Basic & $-0.80{}^{*}$ & --- & $-0.98{}^{*}$ & $-0.92{}^{*}$ & $-0.48{}^{*}$ & $-0.58{}^{*}$ & $-0.96{}^{*}$ \\
E2 & Int: Basic$\to$Mid & $+0.38{}^{*}$ & --- & $+0.12$ & $-0.21$ & $-0.16{}^{*}$ & $-0.06$ & $+0.08$ \\
E3 & Int: Mid$\to$Pro & $-0.86{}^{*}$ & --- & $-0.82{}^{*}$ & $-0.41{}^{*}$ & $-0.46{}^{*}$ & $-0.10{}^{*}$ & $-0.44{}^{*}$ \\
E4 & Int: None$\to$Pro & $-0.98{}^{*}$ & --- & $-0.98{}^{*}$ & $-0.94{}^{*}$ & $-0.76{}^{*}$ & $-0.72{}^{*}$ & $-0.92{}^{*}$ \\
F1 & 20 vs 5 personas & $+0.32{}^{*}$ & $+0.10$ & $+0.22{}^{*}$ & $-0.10$ & $+0.29{}^{*}$ & $+0.24$ & $+0.16{}^{*}$ \\
\bottomrule
\end{tabular}
\end{table}

%% file: tables/T7_pairwise_cc.tex
\begin{table}[!htbp]
\centering
\small
\caption{Pairwise comparisons on CC: matched-pairs Cliff's $\delta$ (paired rank-biserial form). Categorical thresholds follow~\citet{romano2006}: neg ($|\delta|<0.147$), sma ($<0.33$), med ($<0.474$), lar ($\geq 0.474$). Asterisk ($*$) denotes $p_{\textsc{fdr}} < 0.05$ after Benjamini--Hochberg correction. Em-dash indicates GPT-mini did not run that comparison. Positive values indicate that the first-listed condition is higher.}
\label{tab:pair-cc}
\begin{tabular}{llrrrrrrr}
\toprule
ID & Comparison & DeepSeek & GPT-mini & Gemini & Kimi & Llama & Grok & Qwen \\
\midrule
A1 & None$\to$Role & $+0.08$ & $-0.64{}^{*}$ & $-0.80{}^{*}$ & $-0.49{}^{*}$ & $+0.19{}^{*}$ & $-0.26{}^{*}$ & $-0.58{}^{*}$ \\
A2 & Role$\to$Basic & $-0.06$ & $+0.79{}^{*}$ & $+0.95{}^{*}$ & $+0.48{}^{*}$ & $-0.26{}^{*}$ & $+0.05$ & $-0.13$ \\
A3 & Basic$\to$Mid & $-0.06$ & $-0.53{}^{*}$ & $-0.33{}^{*}$ & $-0.33{}^{*}$ & $-0.07$ & $-0.08$ & $-0.26$ \\
A4 & Mid$\to$Pro & $-0.76{}^{*}$ & $-0.38{}^{*}$ & $-0.40{}^{*}$ & $-0.17$ & $-0.19{}^{*}$ & $-0.26{}^{*}$ & $-0.07$ \\
A5 & None$\to$Pro & $-0.32{}^{*}$ & $-0.48{}^{*}$ & $-0.57{}^{*}$ & $-0.49{}^{*}$ & $-0.07$ & $-0.32{}^{*}$ & $-0.64{}^{*}$ \\
B1 & vs Enhanced & $-0.24{}^{*}$ & --- & $-0.61{}^{*}$ & $-0.78{}^{*}$ & $+0.86{}^{*}$ & $-0.51{}^{*}$ & $-0.69{}^{*}$ \\
B2 & vs TempHigh & $-0.21{}^{*}$ & --- & $-0.56{}^{*}$ & --- & $+0.68{}^{*}$ & $-0.62{}^{*}$ & $-0.38{}^{*}$ \\
B3 & vs Dimension & $+0.22$ & --- & $-0.24{}^{*}$ & $-0.18{}^{*}$ & $+0.82{}^{*}$ & $-0.08$ & $-0.43{}^{*}$ \\
B4 & vs RandomSys & $-0.27{}^{*}$ & --- & $-0.96{}^{*}$ & $-0.78{}^{*}$ & $+0.56{}^{*}$ & $-0.53{}^{*}$ & $-0.98{}^{*}$ \\
C1 & Indep$\to$Seq @None & $-0.96{}^{*}$ & --- & $-0.95{}^{*}$ & $-0.82{}^{*}$ & $-0.30{}^{*}$ & $-0.71{}^{*}$ & $-0.90{}^{*}$ \\
C2 & Indep$\to$Int @None & $-0.38{}^{*}$ & --- & $-0.30{}^{*}$ & $-0.96{}^{*}$ & $+0.52{}^{*}$ & $+0.07$ & $-0.78{}^{*}$ \\
C3 & Seq vs Int @None & $+0.94{}^{*}$ & --- & $+0.85{}^{*}$ & $-0.66{}^{*}$ & $+0.88{}^{*}$ & $+0.70{}^{*}$ & $-0.23{}^{*}$ \\
C4 & Indep$\to$Seq @Pro & $-1.00{}^{*}$ & $-1.00{}^{*}$ & $-1.00{}^{*}$ & $-1.00{}^{*}$ & $-1.00{}^{*}$ & $-1.00{}^{*}$ & $-0.96{}^{*}$ \\
C5 & Indep$\to$Int @Pro & $-0.12$ & $-0.21{}^{*}$ & $+0.23{}^{*}$ & $-0.44{}^{*}$ & $+0.60{}^{*}$ & $+0.55{}^{*}$ & $-0.46{}^{*}$ \\
C6 & Seq vs Int @Pro & $+1.00{}^{*}$ & $+1.00{}^{*}$ & $+1.00{}^{*}$ & $+1.00{}^{*}$ & $+1.00{}^{*}$ & $+1.00{}^{*}$ & $+0.96{}^{*}$ \\
C7 & Indep$\to$Seq @Role & $-1.00{}^{*}$ & --- & $-1.00{}^{*}$ & $-1.00{}^{*}$ & $-1.00{}^{*}$ & $-1.00{}^{*}$ & $-1.00{}^{*}$ \\
C8 & Indep$\to$Int @Role & $-0.16$ & --- & $+0.58{}^{*}$ & $-0.70{}^{*}$ & $+0.06$ & $-0.11$ & $-0.70{}^{*}$ \\
C9 & Seq vs Int @Role & $+1.00{}^{*}$ & --- & $+1.00{}^{*}$ & $+1.00{}^{*}$ & $+1.00{}^{*}$ & $+1.00{}^{*}$ & $+1.00{}^{*}$ \\
D0 & Seq: None$\to$Role & $-1.00{}^{*}$ & --- & $-1.00{}^{*}$ & $-1.00{}^{*}$ & $-1.00{}^{*}$ & $-1.00{}^{*}$ & $-1.00{}^{*}$ \\
D1 & Seq: None$\to$Basic & $-0.84{}^{*}$ & --- & $-1.00{}^{*}$ & $-0.98{}^{*}$ & $-0.98{}^{*}$ & $-1.00{}^{*}$ & $-1.00{}^{*}$ \\
D2 & Seq: Basic$\to$Mid & $-0.03$ & --- & $+0.56{}^{*}$ & $-0.23{}^{*}$ & $-0.01$ & $-0.07$ & $-0.36{}^{*}$ \\
D3 & Seq: Mid$\to$Pro & $-0.94{}^{*}$ & --- & $-0.38{}^{*}$ & $+0.31{}^{*}$ & $+0.03$ & $-0.06$ & $-0.19{}^{*}$ \\
D4 & Seq: None$\to$Pro & $-1.00{}^{*}$ & --- & $-1.00{}^{*}$ & $-0.96{}^{*}$ & $-0.96{}^{*}$ & $-1.00{}^{*}$ & $-0.96{}^{*}$ \\
E0 & Int: None$\to$Role & $+0.54{}^{*}$ & --- & $-0.17$ & $+0.43{}^{*}$ & $-0.64{}^{*}$ & $-0.39{}^{*}$ & $-0.06$ \\
E1 & Int: None$\to$Basic & $+0.44{}^{*}$ & --- & $+0.38{}^{*}$ & $+0.73{}^{*}$ & $-0.12$ & $+0.17$ & $+0.38{}^{*}$ \\
E2 & Int: Basic$\to$Mid & $+0.34{}^{*}$ & --- & $-0.23{}^{*}$ & $-0.26{}^{*}$ & $-0.22{}^{*}$ & $-0.05$ & $-0.34{}^{*}$ \\
E3 & Int: Mid$\to$Pro & $-0.81{}^{*}$ & --- & $-0.39{}^{*}$ & $+0.23{}^{*}$ & $-0.28{}^{*}$ & $+0.20$ & $-0.13$ \\
E4 & Int: None$\to$Pro & $+0.03$ & --- & $+0.07$ & $+0.66{}^{*}$ & $-0.51{}^{*}$ & $+0.13$ & $+0.02$ \\
F1 & 20 vs 5 personas & $+1.00{}^{*}$ & $+1.00{}^{*}$ & $+1.00{}^{*}$ & $+1.00{}^{*}$ & $+1.00{}^{*}$ & $+1.00{}^{*}$ & $+1.00{}^{*}$ \\
\bottomrule
\end{tabular}
\end{table}

%% file: tables/T7_pairwise_vs.tex
\begin{table}[!htbp]
\centering
\small
\caption{Pairwise comparisons on VS: matched-pairs Cliff's $\delta$ (paired rank-biserial form). Categorical thresholds follow~\citet{romano2006}: neg ($|\delta|<0.147$), sma ($<0.33$), med ($<0.474$), lar ($\geq 0.474$). Asterisk ($*$) denotes $p_{\textsc{fdr}} < 0.05$ after Benjamini--Hochberg correction. Em-dash indicates GPT-mini did not run that comparison. Positive values indicate that the first-listed condition is higher.}
\label{tab:pair-vs}
\begin{tabular}{llrrrrrrr}
\toprule
ID & Comparison & DeepSeek & GPT-mini & Gemini & Kimi & Llama & Grok & Qwen \\
\midrule
A1 & None$\to$Role & $-0.38{}^{*}$ & $-0.76{}^{*}$ & $-1.00{}^{*}$ & $-0.84{}^{*}$ & $-0.24{}^{*}$ & $-0.54{}^{*}$ & $-0.77{}^{*}$ \\
A2 & Role$\to$Basic & $-0.24{}^{*}$ & $+0.78{}^{*}$ & $+0.88{}^{*}$ & $+0.52{}^{*}$ & $-0.34{}^{*}$ & $+0.02$ & $+0.02$ \\
A3 & Basic$\to$Mid & $+0.14$ & $-0.62{}^{*}$ & $-0.34{}^{*}$ & $-0.24{}^{*}$ & $-0.08$ & $-0.18{}^{*}$ & $-0.20{}^{*}$ \\
A4 & Mid$\to$Pro & $-0.80{}^{*}$ & $-0.36{}^{*}$ & $-0.74{}^{*}$ & $-0.30{}^{*}$ & $-0.42{}^{*}$ & $-0.40{}^{*}$ & $-0.38{}^{*}$ \\
A5 & None$\to$Pro & $-0.86{}^{*}$ & $-0.62{}^{*}$ & $-0.90{}^{*}$ & $-0.82{}^{*}$ & $-0.56{}^{*}$ & $-0.68{}^{*}$ & $-0.92{}^{*}$ \\
B1 & vs Enhanced & $-0.26{}^{*}$ & --- & $-0.48{}^{*}$ & $-0.76{}^{*}$ & $+0.62{}^{*}$ & $-0.42{}^{*}$ & $-0.60{}^{*}$ \\
B2 & vs TempHigh & $-0.22{}^{*}$ & --- & $-0.52{}^{*}$ & --- & $+0.44{}^{*}$ & $-0.58{}^{*}$ & $-0.43{}^{*}$ \\
B3 & vs Dimension & $+0.20$ & --- & $-0.24{}^{*}$ & $-0.31{}^{*}$ & $+0.71{}^{*}$ & $+0.00$ & $-0.52{}^{*}$ \\
B4 & vs RandomSys & $-0.38{}^{*}$ & --- & $-0.96{}^{*}$ & $-0.78{}^{*}$ & $+0.30{}^{*}$ & $-0.48{}^{*}$ & $-0.96{}^{*}$ \\
C1 & Indep$\to$Seq @None & $-0.98{}^{*}$ & --- & $-0.92{}^{*}$ & $-0.94{}^{*}$ & $-0.66{}^{*}$ & $-0.94{}^{*}$ & $-0.88{}^{*}$ \\
C2 & Indep$\to$Int @None & $-0.26{}^{*}$ & --- & $-0.34{}^{*}$ & $-0.94{}^{*}$ & $+0.26{}^{*}$ & $-0.34{}^{*}$ & $-0.88{}^{*}$ \\
C3 & Seq vs Int @None & $+0.96{}^{*}$ & --- & $+0.78{}^{*}$ & $-0.56{}^{*}$ & $+0.78{}^{*}$ & $+0.56{}^{*}$ & $-0.26{}^{*}$ \\
C4 & Indep$\to$Seq @Pro & $-1.00{}^{*}$ & $-1.00{}^{*}$ & $-1.00{}^{*}$ & $-1.00{}^{*}$ & $-1.00{}^{*}$ & $-1.00{}^{*}$ & $-0.96{}^{*}$ \\
C5 & Indep$\to$Int @Pro & $-0.28$ & $-0.40{}^{*}$ & $-0.16{}^{*}$ & $-0.82{}^{*}$ & $+0.28{}^{*}$ & $-0.04$ & $-0.82{}^{*}$ \\
C6 & Seq vs Int @Pro & $+0.98{}^{*}$ & $+0.96{}^{*}$ & $+1.00{}^{*}$ & $+0.94{}^{*}$ & $+1.00{}^{*}$ & $+1.00{}^{*}$ & $+0.82{}^{*}$ \\
C7 & Indep$\to$Seq @Role & $-1.00{}^{*}$ & --- & $-1.00{}^{*}$ & $-1.00{}^{*}$ & $-1.00{}^{*}$ & $-1.00{}^{*}$ & $-1.00{}^{*}$ \\
C8 & Indep$\to$Int @Role & $-0.18$ & --- & $+0.04$ & $-0.84{}^{*}$ & $-0.22{}^{*}$ & $-0.70{}^{*}$ & $-0.92{}^{*}$ \\
C9 & Seq vs Int @Role & $+1.00{}^{*}$ & --- & $+1.00{}^{*}$ & $+1.00{}^{*}$ & $+1.00{}^{*}$ & $+1.00{}^{*}$ & $+0.98{}^{*}$ \\
D0 & Seq: None$\to$Role & $-1.00{}^{*}$ & --- & $-1.00{}^{*}$ & $-1.00{}^{*}$ & $-1.00{}^{*}$ & $-1.00{}^{*}$ & $-1.00{}^{*}$ \\
D1 & Seq: None$\to$Basic & $-0.96{}^{*}$ & --- & $-1.00{}^{*}$ & $-1.00{}^{*}$ & $-0.94{}^{*}$ & $-1.00{}^{*}$ & $-1.00{}^{*}$ \\
D2 & Seq: Basic$\to$Mid & $+0.24{}^{*}$ & --- & $+0.10$ & $-0.36{}^{*}$ & $-0.12$ & $-0.31{}^{*}$ & $-0.56{}^{*}$ \\
D3 & Seq: Mid$\to$Pro & $-0.96{}^{*}$ & --- & $-0.88{}^{*}$ & $-0.29{}^{*}$ & $-0.46{}^{*}$ & $-0.43{}^{*}$ & $-0.64{}^{*}$ \\
D4 & Seq: None$\to$Pro & $-1.00{}^{*}$ & --- & $-1.00{}^{*}$ & $-1.00{}^{*}$ & $-0.96{}^{*}$ & $-1.00{}^{*}$ & $-0.96{}^{*}$ \\
E0 & Int: None$\to$Role & $-0.22{}^{*}$ & --- & $-0.84{}^{*}$ & $-0.52{}^{*}$ & $-0.62{}^{*}$ & $-0.70{}^{*}$ & $-0.80{}^{*}$ \\
E1 & Int: None$\to$Basic & $-0.24{}^{*}$ & --- & $-0.70{}^{*}$ & $+0.07$ & $-0.18{}^{*}$ & $-0.22$ & $-0.46{}^{*}$ \\
E2 & Int: Basic$\to$Mid & $+0.28{}^{*}$ & --- & $-0.04$ & $-0.23{}^{*}$ & $-0.16$ & $-0.02$ & $-0.14$ \\
E3 & Int: Mid$\to$Pro & $-0.86{}^{*}$ & --- & $-0.76{}^{*}$ & $-0.07$ & $-0.30{}^{*}$ & $-0.02$ & $-0.26{}^{*}$ \\
E4 & Int: None$\to$Pro & $-0.82{}^{*}$ & --- & $-0.90{}^{*}$ & $+0.03$ & $-0.58{}^{*}$ & $-0.16{}^{*}$ & $-0.70{}^{*}$ \\
F1 & 20 vs 5 personas & $+1.00{}^{*}$ & $+0.98{}^{*}$ & $+1.00{}^{*}$ & $+1.00{}^{*}$ & $+1.00{}^{*}$ & $+1.00{}^{*}$ & $+1.00{}^{*}$ \\
\bottomrule
\end{tabular}
\end{table}

%% file: tables/T8_beta.tex
\begin{table}[!htbp]
\centering
\footnotesize
\caption{$\beta$-Vendi Score and Unique Cluster Ratio (UCR) for each comparison. UCR cells show ``A-only \% / B-only \%''. Compare against the split-half baseline ($\beta$-VS $\approx 4.0$, UCR $\approx 30\%$) in Table~\ref{tab:baseline}.}
\label{tab:beta}
\resizebox{\textwidth}{!}{\begin{tabular}{lrrrrrrr}
\toprule
Comparison & DeepSeek & GPT-mini & Gemini & Kimi & Llama & Grok & Qwen \\
\midrule
\multicolumn{8}{l}{\textit{$\beta$-Vendi Score}} \\
\quad None vs.\ Role & 7.87 & 4.22 & 10.75 & 9.31 & 4.04 & 7.13 & 7.98 \\
\quad None vs.\ Basic & 8.99 & 3.43 & 10.91 & 9.26 & 5.14 & 7.38 & 8.22 \\
\quad None vs.\ Pro & 9.83 & 4.10 & 13.36 & 10.53 & 5.84 & 8.67 & 9.72 \\
\quad Basic vs.\ Pro & 10.29 & 3.93 & 11.93 & 10.59 & 5.16 & 8.19 & 9.88 \\
\quad Single-Call vs.\ Multi-Turn (5p Pro) & 10.19 & 6.63 & 12.29 & 10.56 & 5.85 & 10.17 & 11.57 \\
\quad Single-Call vs.\ Multi-Agent (5p Pro) & 10.12 & 5.95 & 12.52 & 11.33 & 5.67 & 9.95 & 12.48 \\
\quad Multi-Turn vs.\ Multi-Agent (5p Pro) & 15.15 & 8.26 & 17.30 & 17.94 & 7.14 & 14.73 & 18.30 \\
\quad Single-Call vs.\ Multi-Turn (None) & 6.73 & --- & 7.66 & 8.48 & 4.15 & 8.27 & 6.12 \\
\quad Single-Call vs.\ Multi-Agent (None) & 7.27 & --- & 7.82 & 10.79 & 4.19 & 8.99 & 7.52 \\
\quad Single-Call: Pro vs.\ Mid & 10.00 & --- & 10.80 & 10.80 & 4.65 & 7.43 & 9.28 \\
\quad Single-Call: Pro vs.\ Role & 11.47 & --- & 14.62 & 12.76 & 5.06 & 7.83 & 10.71 \\
\quad Multi-Turn: Pro vs.\ Mid & 11.87 & --- & 10.62 & 14.30 & 5.93 & 8.33 & 8.94 \\
\quad Multi-Turn: Pro vs.\ Role & 15.22 & --- & 15.41 & 18.23 & 8.06 & 9.70 & 12.58 \\
\quad Multi-Agent: Pro vs.\ Mid & 12.18 & --- & 15.05 & 18.02 & 5.76 & 12.89 & 17.57 \\
\quad Multi-Agent: Pro vs.\ Role & --- & --- & 19.09 & 21.38 & 7.07 & 14.86 & 20.14 \\
\midrule
\multicolumn{8}{l}{\textit{Unique Cluster Ratio (A-only / B-only)}} \\
\quad None vs.\ Role & 54\%/69\% & 30\%/42\% & 58\%/75\% & 59\%/72\% & 31\%/46\% & 46\%/52\% & 49\%/66\% \\
\quad None vs.\ Basic & 60\%/76\% & 28\%/30\% & 58\%/73\% & 60\%/70\% & 40\%/57\% & 47\%/52\% & 50\%/66\% \\
\quad None vs.\ Pro & 60\%/78\% & 31\%/38\% & 71\%/85\% & 65\%/76\% & 44\%/65\% & 53\%/61\% & 59\%/76\% \\
\quad Basic vs.\ Pro & 65\%/67\% & 30\%/36\% & 65\%/72\% & 63\%/68\% & 45\%/51\% & 50\%/54\% & 59\%/64\% \\
\quad Single-Call vs.\ Multi-Turn (5p Pro) & 82\%/86\% & 57\%/71\% & 86\%/88\% & 85\%/88\% & 63\%/70\% & 76\%/82\% & 83\%/86\% \\
\quad Single-Call vs.\ Multi-Agent (5p Pro) & 85\%/87\% & 56\%/65\% & 88\%/89\% & 88\%/90\% & 63\%/67\% & 77\%/80\% & 83\%/88\% \\
\quad Multi-Turn vs.\ Multi-Agent (5p Pro) & 85\%/84\% & 67\%/58\% & 87\%/87\% & 89\%/89\% & 67\%/59\% & 82\%/80\% & 85\%/86\% \\
\quad Single-Call vs.\ Multi-Turn (None) & 38\%/65\% & --- & 39\%/65\% & 50\%/72\% & 25\%/46\% & 49\%/67\% & 35\%/57\% \\
\quad Single-Call vs.\ Multi-Agent (None) & 45\%/63\% & --- & 44\%/57\% & 62\%/80\% & 33\%/38\% & 54\%/64\% & 43\%/67\% \\
\quad Single-Call: Pro vs.\ Mid & 64\%/62\% & --- & 67\%/63\% & 65\%/62\% & 48\%/42\% & 50\%/47\% & 61\%/58\% \\
\quad Single-Call: Pro vs.\ Role & 74\%/70\% & --- & 79\%/75\% & 72\%/72\% & 54\%/41\% & 52\%/49\% & 70\%/65\% \\
\quad Multi-Turn: Pro vs.\ Mid & 73\%/71\% & --- & 72\%/69\% & 76\%/75\% & 52\%/47\% & 61\%/60\% & 67\%/65\% \\
\quad Multi-Turn: Pro vs.\ Role & 81\%/78\% & --- & 83\%/79\% & 82\%/83\% & 59\%/53\% & 64\%/63\% & 75\%/75\% \\
\quad Multi-Agent: Pro vs.\ Mid & 80\%/76\% & --- & 85\%/83\% & 88\%/87\% & 55\%/48\% & 75\%/75\% & 84\%/83\% \\
\quad Multi-Agent: Pro vs.\ Role & --- & --- & 92\%/89\% & 92\%/92\% & 60\%/60\% & 79\%/79\% & 88\%/87\% \\
\bottomrule
\end{tabular}}
\end{table}

%% file: tables/T9_lowcost.tex
\begin{table}[!htbp]
\centering
\small
\caption{$\Delta$MPD relative to the Single-Call $\times$ None baseline for the four low-cost tricks and the simplest persona condition (Single-Call $\times$ Role).}
\label{tab:lowcost}
\begin{tabular}{lrrrrrrr}
\toprule
Condition & DeepSeek & GPT-mini & Gemini & Kimi & Llama & Grok & Qwen \\
\midrule
Enhanced Prompt & +0.0075 & --- & +0.0015 & +0.0150 & +0.0010 & +0.0072 & +0.0125 \\
High Temperature & +0.0023 & --- & +0.0014 & --- & +0.0041 & +0.0009 & +0.0037 \\
Demographic Cueing & -0.0030 & --- & +0.0010 & +0.0036 & +0.0013 & -0.0005 & +0.0107 \\
Trait Assignment & +0.0206 & --- & +0.0385 & +0.0401 & +0.0070 & +0.0100 & +0.0346 \\
Role (reference) & +0.0682 & +0.0328 & +0.0844 & +0.0849 & +0.0410 & +0.0279 & +0.0764 \\
\bottomrule
\end{tabular}
\end{table}

%% file: tables/T12_perturn.tex
\begin{table}[!htbp]
\centering
\footnotesize
\caption{Per-turn cumulative MPD for Multi-Turn and Multi-Agent across two persona depths (None, Pro). Step-1 share is $\text{MPD}_1 / \text{MPD}_3$. Marginal MPD gain is computed on the merged opinion pool through that step.}
\label{tab:perturn}
\begin{tabular}{llrrrr}
\toprule
Condition & Model & Final MPD & Step-1 share & Step-2 marginal & Step-3 marginal \\
\midrule
Multi-Turn $\times$ None & DeepSeek & 0.6584 & 90.6\% & +0.0463 & +0.0159 \\
 & Gemini & 0.6727 & 93.7\% & +0.0292 & +0.0128 \\
 & Kimi & 0.6852 & 91.1\% & +0.0453 & +0.0154 \\
 & Llama & 0.5830 & 91.4\% & +0.0350 & +0.0155 \\
 & Grok & 0.6808 & 90.4\% & +0.0492 & +0.0163 \\
 & Qwen & 0.6525 & 93.3\% & +0.0342 & +0.0094 \\
\midrule
Multi-Turn $\times$ Pro & DeepSeek & 0.7429 & 94.7\% & +0.0304 & +0.0093 \\
 & Gemini & 0.7464 & 96.8\% & +0.0183 & +0.0057 \\
 & Kimi & 0.7519 & 93.5\% & +0.0358 & +0.0131 \\
 & Llama & 0.6682 & 91.8\% & +0.0404 & +0.0143 \\
 & Grok & 0.7248 & 91.6\% & +0.0460 & +0.0152 \\
 & Qwen & 0.7474 & 95.9\% & +0.0246 & +0.0064 \\
\midrule
Multi-Agent $\times$ None & DeepSeek & 0.6144 & 92.0\% & +0.0213 & +0.0280 \\
 & Gemini & 0.6497 & 96.3\% & +0.0133 & +0.0105 \\
 & Kimi & 0.6865 & 92.3\% & +0.0344 & +0.0184 \\
 & Llama & 0.5586 & 95.1\% & +0.0086 & +0.0188 \\
 & Grok & 0.6673 & 94.4\% & +0.0230 & +0.0147 \\
 & Qwen & 0.6727 & 88.9\% & +0.0373 & +0.0374 \\
\midrule
Multi-Agent $\times$ Pro & DeepSeek & 0.7166 & 94.0\% & +0.0274 & +0.0156 \\
 & Gemini & 0.7414 & 95.8\% & +0.0208 & +0.0106 \\
 & Kimi & 0.7551 & 93.2\% & +0.0324 & +0.0193 \\
 & Llama & 0.6277 & 96.2\% & +0.0095 & +0.0144 \\
 & Grok & 0.7069 & 94.5\% & +0.0260 & +0.0126 \\
 & Qwen & 0.7552 & 90.2\% & +0.0454 & +0.0283 \\
\bottomrule
\end{tabular}
\end{table}

%% file: tables/T13_splitgroup.tex
\begin{table}[!htbp]
\centering
\small
\caption{Persona pool breadth: 20 Pro-level personas split into 4 random groups of 5; mean $\beta$-VS / UCR over the 6 group pairs (Independent $\times$ Pro). Reference row gives the cross-depth Basic vs.\ Pro $\beta$-VS for comparison. GPT-mini excluded (5-persona Pro condition not run).}
\label{tab:splitgroup}
\begin{tabular}{lrrrrrr}
\toprule
Quantity & DeepSeek & Gemini & Grok & Kimi & Llama & Qwen \\
\midrule
Mean $\beta$-VS (subgroup pairs) & 8.95 & 10.85 & 7.89 & 8.68 & 4.96 & 9.35 \\
Mean UCR (subgroup pairs) & 77\% & 82\% & 66\% & 79\% & 59\% & 75\% \\
\midrule
Reference: $\beta$-VS Basic vs.\ Pro & 10.29 & 11.93 & 8.19 & 10.59 & 5.16 & 9.88 \\
\bottomrule
\end{tabular}
\end{table}

%% file: tables/T17_density.tex
\begin{table}[!htbp]
\centering
\footnotesize
\caption{Mean number of extracted opinions per question, by condition and model. The decline under persona conditioning is offset by greater dispersion (see Section~\ref{sec:persona-depth}). GPT-5.4-mini is omitted from this table because it ran on only $9$ of the 19 primary conditions plus 2 ablation variants (Appendix~\ref{app:models}); the cells it does cover appear in Table~\ref{tab:alpha-cc-raw}.}
\label{tab:density}
\begin{tabular}{lrrrrrr}
\toprule
Condition & DeepSeek & Gemini & Grok & Kimi & Llama & Qwen \\
\midrule
\texttt{independent-none} & 328 & 382 & 303 & 230 & 422 & 364 \\
\texttt{independent-role} & 152 & 259 & 258 & 162 & 164 & 236 \\
\texttt{independent-basic} & 160 & 232 & 270 & 163 & 164 & 269 \\
\texttt{independent-mid} & 153 & 221 & 259 & 168 & 165 & 268 \\
\texttt{independent-pro} & 188 & 217 & 254 & 160 & 156 & 249 \\
\texttt{independent-enhanced} & 343 & 412 & 330 & 257 & 242 & 427 \\
\texttt{independent-temp-high} & 329 & 382 & 412 & --- & 241 & 362 \\
\texttt{independent-dimension} & 320 & 374 & 306 & 236 & 239 & 403 \\
\texttt{independent-random-sys} & 298 & 349 & 307 & 213 & 228 & 544 \\
\texttt{independent-pro-5} & 48 & 57 & 64 & 41 & 41 & 66 \\
\texttt{sequential-none} & 317 & 307 & 243 & 195 & 193 & 317 \\
\texttt{sequential-role} & 995 & 881 & 947 & 653 & 594 & 1067 \\
\texttt{sequential-basic} & 551 & 1111 & 859 & 519 & 486 & 932 \\
\texttt{sequential-mid} & 542 & 747 & 837 & 524 & 472 & 943 \\
\texttt{sequential-pro} & 654 & 731 & 817 & 486 & 415 & 1299 \\
\texttt{interactive-none} & 227 & 266 & 219 & 335 & 164 & 346 \\
\texttt{interactive-role} & 139 & 189 & 200 & 199 & 142 & 265 \\
\texttt{interactive-basic} & 128 & 158 & 173 & 154 & 121 & 226 \\
\texttt{interactive-mid} & 120 & 164 & 171 & 166 & 132 & 256 \\
\texttt{interactive-pro} & 156 & 171 & 162 & 156 & 127 & 253 \\
\bottomrule
\end{tabular}
\end{table}

%% file: tables/T18_nonerole.tex
\begin{table}[!htbp]
\centering
\small
\caption{$\Delta$MPD from None to Role across architectures. Gains are of similar magnitude across architectures (per-model spread $\leq 0.032$); cf.\ Section~\ref{sec:interaction}.}
\label{tab:nonerole}
\begin{tabular}{lrrrrrrr}
\toprule
Architecture & DeepSeek & GPT-mini & Gemini & Kimi & Llama & Grok & Qwen \\
\midrule
Single-Call & +0.0681 & +0.0328 & +0.0844 & +0.0849 & +0.0411 & +0.0278 & +0.0764 \\
Multi-Turn & +0.0609 & --- & +0.0690 & +0.0836 & +0.0729 & +0.0412 & +0.0942 \\
Multi-Agent & +0.0619 & --- & +0.0832 & +0.0715 & +0.0576 & +0.0497 & +0.0813 \\
\bottomrule
\end{tabular}
\end{table}

%% file: tables/T20_human_eval.tex
\begin{table}[!htbp]
\centering
\small
\caption{Blinded human evaluation of viewpoint breadth (Prolific; 28 sessions, 27 annotators; each item rated twice). Annotators chose which of two complete responses (one generated with a Role persona and one with the corresponding Basic persona) contains a broader range of substantively different viewpoints. $D = P(\text{Basic broader}) - P(\text{Role broader})$, with ties coded as zero; the CI resamples the 42 matched items within model. Agreement is exact agreement between the two ratings of an item and nominal Krippendorff's $\alpha$.}
\label{tab:human-eval}
\resizebox{\linewidth}{!}{%
\begin{tabular}{lrrrrrrr}
\toprule
Comparison & $N$ & Role & Basic & Same & Unable & $D$ [95\% CI] & Agreement ($\alpha$) \\
\midrule
Role--Basic & 84 & 32 (38.1\%) & 24 (28.6\%) & 27 (32.1\%) & 1 & $-0.095$ [$-0.286$, $0.095$] & 52.4\% (0.298) \\
\bottomrule
\end{tabular}%
}
\end{table}

%% file: tables/T21_robustness.tex
\begin{table}[!htbp]
\centering
\small
\caption{Stability of the Single-Call persona-depth results within the 100-question benchmark. $\Delta$MPD values are paired over questions with 95\% task-bootstrap percentile CIs (20{,}000 resamples). ``Initial $-$ extra'' is the None$\to$Role gain minus the Role$\to$Pro gain. The last two columns give the fraction of 10{,}000 random 50-question subsets in which the None$\to$Role gain is positive and in which it exceeds the Role$\to$Pro gain. $N$ is the number of questions with complete data.}
\label{tab:robustness}
\resizebox{\linewidth}{!}{%
\begin{tabular}{lrrrrrr}
\toprule
Model & $N$ & None$\to$Role & Role$\to$Pro & Initial $-$ extra & P(initial$>$0) & P(initial$>$extra) \\
\midrule
DeepSeek & 100 & 0.0681 [0.0567, 0.0802] & 0.0361 [0.0271, 0.0445] & 0.0320 [0.0154, 0.0494] & 100\% & 100\% \\
GPT-mini & 100 & 0.0328 [0.0250, 0.0411] & $-0.0135$ [$-0.0202$, $-0.0075$] & 0.0463 [0.0342, 0.0599] & 100\% & 100\% \\
Gemini & 100 & 0.0844 [0.0767, 0.0924] & 0.0090 [0.0036, 0.0145] & 0.0754 [0.0653, 0.0855] & 100\% & 100\% \\
Kimi & 99 & 0.0843 [0.0760, 0.0928] & $-0.0027$ [$-0.0083$, 0.0029] & 0.0869 [0.0761, 0.0981] & 100\% & 100\% \\
Llama & 100 & 0.0411 [0.0321, 0.0509] & 0.0393 [0.0306, 0.0477] & CI includes 0 & 100\% & 60.7\% \\
Grok & 100 & 0.0278 [0.0211, 0.0353] & 0.0114 [0.0054, 0.0167] & 0.0165 [0.0059, 0.0280] & 100\% & 100\% \\
Qwen & 97 & 0.0747 [0.0650, 0.0843] & 0.0140 [0.0067, 0.0212] & 0.0607 [0.0467, 0.0748] & 100\% & 100\% \\
\bottomrule
\end{tabular}%
}
\end{table}

%% file: tables/T22_ma_prompt.tex
\begin{table}[!htbp]
\centering
\small
\caption{Multi-Agent prompt sensitivity (DeepSeek-V4-Flash, 100 questions, five fixed Pro personas, three rounds). Round~1 is generated once and shared; only the Round~2--3 instruction differs. Values are computed on Rounds~2--3, with CC and VS jointly rarefied across the four conditions. ``Opinions'' is the mean number of extracted opinions per question.}
\label{tab:ma-prompt}
\begin{tabular}{lrrrr}
\toprule
Round 2--3 instruction & Opinions & MPD & CC (rarefied) & VS (rarefied) \\
\midrule
Neutral (main study) & 71.7 & \textbf{0.7230} & \textbf{57.30} & \textbf{26.68} \\
Agreement & 94.6 & 0.6898 & 45.90 & 21.74 \\
Disagreement & 120.4 & 0.7160 & 56.88 & 26.00 \\
Coverage & 79.6 & 0.6797 & 57.09 & 23.40 \\
\bottomrule
\end{tabular}
\end{table}

%% file: tables/T16_compute.tex
\begin{table}[!htbp]
\centering
\small
\caption{Compute resources for the full experiment, aggregated from per-condition runtime logs. ``Conditions logged'' is the number of primary conditions and ablation variants for which a complete log was located (out of the 19 primary conditions plus 2 ablation variants run); records and tokens are summed over those conditions. The full experiment used commercial chat-completion APIs only; no GPU compute was required. Per-model dollar cost is not broken out because the seven models use heterogeneous input/output pricing schedules through OpenRouter and direct APIs that changed during the experiment window; the aggregate spend is reported in the surrounding text.}
\label{tab:compute}
\begin{tabular}{lrrr}
\toprule
Model & Conditions logged & Total records & Total tokens \\
\midrule
DeepSeek & 20 & 62,515 & 74.5\,M \\
GPT-mini & 7  & 17,500 & 15.3\,M \\
Gemini   & 18 & 46,760 & 66.1\,M \\
Kimi     & 14 & 37,500 & 27.9\,M \\
Llama    & 18 & 45,620 & 53.8\,M \\
Grok     & 18 & 44,880 & 85.0\,M \\
Qwen     & 18 & 46,634 & 92.7\,M \\
\midrule
\textbf{Total} & --- & \textbf{301,409} & \textbf{415\,M} \\
\bottomrule
\end{tabular}
\end{table}